\documentclass[nohyperref, accepted]{article}

\usepackage{subfiles}
\usepackage{main_arxiv/custom_style}

\usepackage{microtype}
\usepackage{graphicx}
\usepackage{subfigure}
\usepackage{booktabs} 

\usepackage[bookmarks=false]{hyperref}

\usepackage{main_arxiv/icml2022}

\usepackage{amsmath}
\usepackage{amssymb}
\usepackage{mathtools}
\usepackage{amsthm}

\usepackage[capitalize,noabbrev]{cleveref}

\theoremstyle{plain}
\newtheorem{theorem}{Theorem}[section]
\newtheorem{proposition}[theorem]{Proposition}
\newtheorem{lemma}[theorem]{Lemma}
\newtheorem{corollary}[theorem]{Corollary}
\theoremstyle{definition}
\newtheorem{definition}[theorem]{Definition}

\theoremstyle{remark}
\newtheorem{remark}[theorem]{Remark}
\newtheorem{example}[theorem]{Example}

\usepackage[textsize=tiny]{todonotes}

\icmltitlerunning{Group invariant machine learning by fundamental domain projections}

\begin{document}

\twocolumn[
\icmltitle{Group invariant machine learning by fundamental domain projections}



\icmlsetsymbol{equal}{*}

\begin{icmlauthorlist}
\icmlauthor{Benjamin Aslan}{equal,ucl}
\icmlauthor{Daniel Platt}{equal,kcl}
\icmlauthor{David Sheard}{equal,ucl}
\end{icmlauthorlist}

\icmlaffiliation{ucl}{Department of Mathematics, University College London, London, UK}
\icmlaffiliation{kcl}{Department of Mathematics, King's College London, London, UK}

\icmlcorrespondingauthor{Benjamin Aslan}{benjamin.aslan.17@ucl.ac.uk}
\icmlcorrespondingauthor{Daniel Platt}{daniel.platt@kcl.ac.uk}
\icmlcorrespondingauthor{David Sheard}{david.sheard.17@ucl.ac.uk}

\icmlkeywords{Group action, Symmetry, Invariant machine learning, Equivariant machine learning, Supervised learning, Geometric topology, Geometric deep learning}

\vskip 0.3in
]



\printAffiliationsAndNotice{\icmlEqualContribution} 

\begin{abstract}
We approach the well-studied problem of supervised group invariant and equivariant machine learning from the point of view of geometric topology. We propose a novel approach using a pre-processing step, which involves projecting the input data into a geometric space which parametrises the orbits of the symmetry group. This new data can then be the input for an arbitrary machine learning model (neural network, random forest, support-vector machine etc). 

We give an algorithm to compute the geometric projection, which is efficient to implement, and we illustrate our approach on some example machine learning problems (including the well-studied problem of predicting Hodge numbers of CICY matrices), in each case finding an improvement in accuracy versus others in the literature. The geometric topology viewpoint also allows us to give a unified description of so-called \emph{intrinsic} approaches to group equivariant machine learning, which encompasses many other approaches in the literature.
\end{abstract}

	\section{Introduction}\label{sec:introduction}

	Many tasks in machine learning can be understood as approximating a function $\al\colon X \to Y$ from a subset $X\subset \R^n$, to a (possibly discrete) subset $Y$ of $\R^m$.
	We consider the problem 
	in the presence of \emph{symmetries}.
	More precisely, suppose a group $G$ that acts on $\R^n$ on the left, and we assume the function $\al$ satisfies the invariance property
	\begin{align}
	\label{equation:invariance-property}
	 \al(g\cdot x)=\al(x)
	 \text{ for all }
	 x \in \R^n, g \in G.
	\end{align}
	In this paper we focus on subgroups of the symmetric group $S_n$ acting on the input space $\R^n$ by permuting the coordinates. A simple example of this type of problem is recognising a single handwritten digit which may have been rotated by $90^\circ$, $180^\circ$, or $270^\circ$, in which case the problem is invariant under the action of the group $\Z_4$.
	
	Machine learning techniques such as neural networks or random forests can be used to approximate $\al$, but the resulting function $\beta$ will, in general, not be invariant under the action $G$.
	The key task is to define machine learning algorithms producing functions $\be\colon \R^n \rightarrow \R^m$ which are guaranteed to satisfy the invariance property \cref{equation:invariance-property}.
	
	\subsection{Previous work}\label{subsec:lit-review}
	Machine learning models which are invariant (or equivariant) under the action of a group $G$ have been extensively studied in the machine learning literature. 
	In \cite{Yarotsky2021}, Yarotsky distinguishes two different approaches to the problem: \emph{symmetrisation based} and \emph{intrinsic} approaches. The first involves averaging some non $G$-invariant model over the action of $G$ to produce an (approximately) $G$-invariant model; whereas intrinsic approaches involve designing the model to be $G$-invariant \emph{a priori} by imposing conditions coming from the group action. 
	
	A standard approach to the problem falling into the first category is \emph{data augmentation}, which has been used in early works such as \cite{Krizhevsky2012}, and is surveyed in \cite{Chen2020}. It involves expanding the training data $D_\textrm{train}=\{(x,y)\mid x\in X_\textrm{train}\subset X, y=\al(x)\in Y\}$ by applying sample elements $G_0\subset G$ to the input. The new training data is then 
	\[
	D^\textrm{aug}_\textrm{train}\coloneqq\{(g\cdot x,y)\mid (x,y)\in D_\textrm{train}\;\textrm{and}\; g\in G_0\}.
	\]
	A similar approach is to take a machine learning architecture $\be$ and apply it to several $G$-translates of an input, before applying a pooling map to these different outputs. This yields a $G$-invariant map, and was studied in \cite{Bao2019}.
	
%
%
	
	We now turn to examples of \emph{intrinsic} approaches.
	For neural networks, one can impose restrictions on the weights so that the resulting network is invariant under a group action on the input.
	This was done, for example, in \cite{Hartford2018} and \cite{Zaheer2017}.
	The same idea is also used in \cite{Maron2020,Ravanbakhsh2016, Ravanbakhsh2017}. Another intrinsic approach is proposed in Section 2 of \cite{Yarotsky2021} based on the theory of \emph{polynomial invariants} of $G$.

	\subsection{Our contribution}\label{sec:our-contribution}
	
	Our approach to the problem is intrinsic, based on the fact that composing a $G$-invariant map with any other map, results in a $G$-invariant map. More precisely, we suggest a \emph{$G$-invariant pre-processing step} to be applied to the input data that can then be composed with any machine learning architecture. 
	The resulting composition is a $G$-invariant architecture.
	
\begin{figure}[htbp]
		\centering
		\begin{tikzpicture}[scale=2.5]	
			\clip (-.01,-.01) rectangle (3.21,1.01);
			
			\draw[rounded corners=.5, thick] (0,0) rectangle (1,1);
			
			\foreach \x/\y in {45/59, 45/19, 40/48, 52/4, 43/82, 35/38, 25/27, 34/12, 25/79, 19/15, 18/85, 55/38, 79/36, 55/64, 32/7}{
				\fill[magenta!70] (\x/100,\y/100) circle (.5pt);
			}
			
			\node at (1,0)[anchor=south east] {$D_\textrm{train}$};
			
			\begin{scope}[xshift=1.65cm]
				
				\begin{scope}[xshift=-0.55cm]
					
					\draw[rounded corners=.5, thick] (0,0) rectangle (1,1);
					
					\foreach \x/\y in {45/59, 45/19, 40/48, 52/4, 43/82, 35/38, 25/27, 34/12, 25/79, 19/15, 18/85, 55/38, 79/36, 55/64, 32/7}{
						\fill[magenta!70] (\x/100,\y/100) circle (.5pt);
						\fill[magenta!50!black, rotate around={120:(.5,.5)}] (\x/100,\y/100) circle (.5pt);			
						\fill[magenta!30, rotate around={-120:(.5,.5)}] (\x/100,\y/100) circle (.5pt);
					}
					
					\node at (1,0)[anchor=south east] {$D_\textrm{train}^\textrm{aug}$};
				\end{scope}
				
				\begin{scope}[xshift=0.55cm]			
					\begin{scope}
						\clip (0,0) rectangle (1,1);
						\fill[cyan!20] (1,1.366) -- (0.5,0.5) -- (1,-.366) -- cycle;
						\foreach \x in {0,120,240}{
							\draw[cyan, dashed, rotate around={\x:(.5,.5)}] (1,1.366) -- (0.5,0.5);
						}
					\end{scope}
					\draw[rounded corners=.5, thick] (0,0) rectangle (1,1);		
					
					\foreach \x/\y in {45/59, 45/19, 40/48, 52/4, 43/82, 35/38, 25/27, 34/12, 25/79, 19/15, 18/85, 55/38, 79/36, 55/64, 32/7}{
						\draw[magenta!40, thin] (\x/100,\y/100) circle (.5pt);
					}
					
					\begin{scope}
						\clip (1,1.366) -- (0.5,0.5) -- (1,-.366) -- cycle;
						
						\foreach \x/\y in {45/59, 45/19, 40/48, 52/4, 43/82, 35/38, 25/27, 34/12, 25/79, 19/15, 18/85, 55/38, 79/36, 55/64, 32/7}{
							\fill[magenta!70] (\x/100,\y/100) circle (.5pt);
							\fill[magenta!70, rotate around={120:(.5,.5)}] (\x/100,\y/100) circle (.5pt);			
							\fill[magenta!70, rotate around={-120:(.5,.5)}] (\x/100,\y/100) circle (.5pt);
						}
					\end{scope}
					
					\node at (1,1)[anchor=north east, cyan!70!black] {\footnotesize$\fd$};
					
					\node at (1,0)[anchor=south east] {$D^\pi_\textrm{train}$};
				\end{scope}
				
			\end{scope}
		\end{tikzpicture}
		\caption{Example training data $D_\textrm{train}$ for a problem invariant under rotations of $120^\circ$ (left). Also the processed training data after augmentation $D_\textrm{train}^\textrm{aug}$, and our approach $D^\pi_\textrm{train}$ (mapping all the data to the blue subset, a fundamental domain for the action).}
		\label{fig:preprocessing}
	\end{figure}
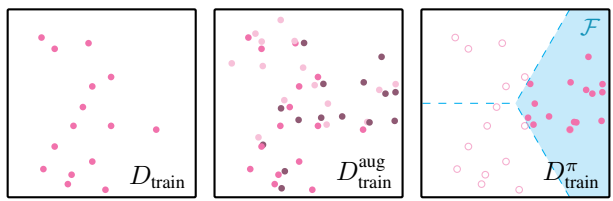

	One way of getting a $G$-invariant self-map of the feature space is to map to a so-called \emph{fundamental domain} $\fd$, which preserves the local geometry of the feature space. The set $\fd \subset X$  comes with a $G$-invariant map $\pi \colon X \rightarrow \fdc$ onto its closure. Let $\overline{\alpha}$ be the restriction of $\alpha$ to $\fdc$, then by $G$-invariance $\alpha=\overline{\alpha}\circ\pi$.
	Instead of fitting a machine learning model $\beta \colon X \rightarrow Y$ to the training data ${D_{\mathrm{train}}\subset X\times Y}$, which approximates $\alpha$, we train the model $\overline{\beta}\colon\fdc\to Y$ with 
	$D^\pi_{\mathrm{train}}\coloneqq\{(\pi(x),y)\mid (x,y)\in D_{{\mathrm{train}}}\}\subset \fdc \times Y$ which approximates $\overline{\alpha}$. 
	The resulting map $\beta=\overline{\beta} \circ \pi \colon X \rightarrow Y$ is $G$-invariant because of the $G$-invariance of $\pi$. The difference between the pre-processing approaches of augmentation and our method is illustrated in \cref{fig:preprocessing}.
	This approach also extends easily to $G$-\emph{equivariant} machine learning, this is explained at the end of \cref{subsec:fun-dom}.
	
    More precisely, $G$-invariant maps from $\R^n$ are pa\-ram\-e\-tris\-ed by maps from the so-called \emph{quotient space} $\R^n/G$ of $\R^n$, see \cref{sec:unification}. The set $\fdc\subset\R^n$ locally models $\R^n/G$, and has the advantage of being extremely easy to compute. 
    Another advantage our approach has is that it can be applied directly to any supervised machine learning model, not just neural networks.

	The rest of the paper is organised as follows. \Cref{sec:mathematical-approach} describes our approach in detail, and compares it with other approaches from the literature. \Cref{sec:results} discusses several applications of our approach, and compares the accuracies of machine learning architectures employing different approaches to $G$-invariant machine learning. Our main application is the learning of Hodge numbers from CICY matrices, first studied in \cite{He2017}, and our method improves on the state of the art \cite{Erbin2021}.
	In \cref{sec:unification} we give a unified view of intrinsic approaches to $G$-equivariant machine learning and describe how each method fits into this framework.

	\subsection*{Acknowledgements}
	We would like to thank Yang-Hui He and Momchil Konstantinov for their comments on a draft of this paper. All three authors were supported by the Engineering and Physical Sciences Research Council [EP/L015234/1]. The EPSRC Centre for Doctoral Training in Geometry and Number Theory (The London School of Geometry and Number Theory), University College London. The second author was supported by the Simons Collaboration ``Special Holonomy in Geometry, Analysis, and Physics'' during part of the work.

	\section{Mathematical approach}\label{sec:mathematical-approach}
	
	In principle, our approach works for a large class of group actions, however in this paper we focus on the case that $G$ is a subgroup of the permutation group $S_n$. Then $G$ acts on $\R^n$ by maps which permute coordinates. 
	More precisely, if $x=(x_i)_i\in\R^n$ and $s\in S_n$ we say $s$ acts on the left by $s\cdot (x_i)_i=\left (x_{s^{-1}(i)}\right )_i$. This induces an action of $G$ on $\R^n$ which we call a \textit{permutation action of $G$}. 
		
	The permutation action of $G$ on $\R^n$ is by isometries in the sense that the Euclidean metric is invariant. For this reason, the action respects the geometry of $\R^n$, and we will exploit this geometry.
	
	\subsection{Fundamental domains}\label{subsec:fun-dom}
	
	Given $x\in \R^n$, its \textit{orbit} under the action of $G$ is the set ${G\cdot x} =\{g\cdot x\mid g\in G\}$. We want to approximate a $G$-invariant function $\al$ which satisfies \cref{equation:invariance-property}. It follows from this equation that $\al$ takes the same value on every element of a  $G$-orbit. A set $R\subset \R^n$ is a \textit{set of orbit representatives} if for all $x\in\R^n$, $R\cap(G\cdot x)\neq\emptyset$. If we approximate $\al$ on a set of orbit representatives then we have essentially approximated it everywhere. A nice choice of orbit representatives which takes into account the geometry of the group action is given by a fundamental domain.
	
	\begin{definition}\label{def:fund-dom}
		Let a finite group $G$ act on $\R^n$ by permuting coordinates. A subset $\fd\subset \R^n$ is called a \textit{fundamental domain} for $G$ if
		\begin{enumerate}\itemsep0em
			\item\label{point:open} The set $\fd$ is open and connected
			\item\label{point:cover} Every $G$-orbit intersects $\fdc$, the closure of $\fd$, in at least one point
			\item\label{point:nonintersecting} If a $G$-orbit intersects $\fdc$ at a point in $\fd$, then this is the unique point of intersection with $\fdc$
		\end{enumerate}
	\end{definition}
	\begin{example}
		If $\Z_2$ acts on $\R^2$ by $(x,y)\mapsto(-x,-y)$, then $\{(x,y)\mid x>0\}$ is a fundamental domain.
	\end{example}
	
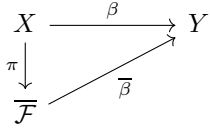
\begin{figure}[htbp]
		\centering
		\begin{tikzcd}
			X \arrow[rr,swap, "\be"'] \arrow[d,swap, "\pi"] &  & Y \\
			\overline{\mathcal{F}} \arrow[rru,swap, "\overline{\be}"]           &  &  
		\end{tikzcd}
		\caption{The role of the projection onto the fundamental domain $\pi$ in constructing $G$-invariant maps.}
		\label{figure:beta-diagram}
	\end{figure}

	Given $G$ acting on $\R^n$ we will find a $G$-invariant map $\pi\colon \R^n\to \fdc$, defined as $\pi(x)= \phi(x)\cdot x$, where ${\phi\colon X\to G}$ is some suitable function. We call such a map a \textit{projection onto the fundamental domain $\fd$}. We can now apply any machine leaning architecture to approximate the function $\restrict{\al}{\fdc}\colon \fdc\to \R^m$ trained on the data $D^\pi_{\mathrm{train}}$ yielding a function $\overline{\be}$. This can then be used to compute the $G$-invariant approximation for $\al$ defined on the whole of $X$ by defining $\be=\overline{\be}\circ\pi$, see \cref{figure:beta-diagram}.
	
	If $\overline{\beta}$ is defined using a neural network, then the natural universal approximation property is satisfied, namely $\beta$ can approximate any continuous, $G$-invariant map $\alpha$ arbitrarily closely.
	This is a direct consequence of the standard universal approximation theorem for neural networks and is proved in \cref{subsec:univ-approx-thm}.
	
	For generic $x\in X$ and any $g\in G$, $\phi$ has the property
	\begin{equation}\label{eq:phi}
		\phi(g\cdot x)=\phi(x)g^{-1}
	\end{equation}
	from which we can show that $\beta$ is indeed $G$-invariant:
	\begin{align*}
	\beta(g\cdot x)&=\overline{\beta}(\phi(g\cdot x)\cdot (g\cdot x))=\overline{\beta}((\phi(x)g^{-1})\cdot (g\cdot x))\\
	&=\overline{\beta}((\phi(x)\cdot x)=\beta(x).
	\end{align*}

	Our method of producing a $G$-invariant architecture can easily be modified to the $G$-equivariant setting. Let $G$ act on $X$ and $Y$, then a $G$-equivariant map $\alpha\colon X\to Y$ satisfies $\alpha(g\cdot x)=g\cdot\alpha(x)$ for all $x\in X$ and $g\in G$. Let $\pi\colon X\to\fdc$ be the fundamental domain projection as above, and define $\phi \colon X\to G$ be a function such that $\pi(x)=\phi(x)\cdot x$.
	Then we define the $\beta$ model via
	\[
	\be(x)\coloneqq\phi(x)^{-1}\cdot\overline{\be}(\pi(x)) =\phi(x)^{-1}\cdot\overline{\be}(\phi(x)\cdot x).
	\]
	As above, for a generic $x\in X$ \cref{eq:phi} implies that $\beta$ is $G$-equivariant.
	
	\begin{remark}\label{remark:boundary}
		The projection $\pi$ is continuous on the preimage of $\fd$ in $\R^n$, but may fail to be continuous on $\partial\fd$, the preimage of the boundary of $\fd$. For an $x$ which is mapped to $\partial\fd$, $\phi$ does not necessarily satisfy \cref{eq:phi}, and so the function $\be$ may not be strictly $G$-invariant/equivariant on $\partial\fd$. This only presents a problem if a significant portion of $D^\pi_{\mathrm{train}}$ lies in $\partial\fd$. We 
		give an example where this problem arises in \cref{sec:results}.
	\end{remark}
	
	In the next two sections we will describe two methods of finding a fundamental domain projection for a given action.
	
	\subsection{Combinatorial projections}
	\label{subsec:comb-projections}
	
	The first projection onto a fundamental domain is combinatorial in nature. For simple examples of group actions it is straightforward to find a projection onto a fundamental domain. We will give two such examples, and then give an algorithm which can be applied to any group action. 
	
	\begin{example}\label{eg:S_n}
		Let $G=S_n$ be the full permutation group acting on $\R^n$. Given a point $x=(x_1,\dots,x_n)\in \R^n$ we can reorder the entries in any way by elements of $S_n$, and a fundamental domain corresponds to a consistent choice of reordering. One consistent choice would be to have the entries in increasing order, so 
		\[
		\fd=\{(x_1,\dots,x_n)\in \R^n\mid x_i<x_{i+1}\;\textrm{for all}\;i<n\}.
		\]
		The projection $\pi\colon\R^n\to\fdc$ can be easily implemented using any sorting algorithm.
		
		Another example of a simple group action is when ${G=\Z_n<S_n}$ acts by cyclically permuting the coordinates. In this case we can make a consistent choice by ensuring that the first entry is the smallest so
		\[
		\fd=\{(x_1,\dots,x_n)\in \R^n\mid x_1<x_{i}\;\textrm{for all}\;i>1\}.
		\]
	\end{example}
	
	The algorithm to find a fundamental domain projection in general is based on \cite{Dixon1988} in which the authors give an efficient algorithm to find a set of unique coset representatives for an arbitrary subgroup $G\le S_n$. A set of coset representatives can be turned into a set of orbit representatives for the permutation action of $G$ on $\R^n$. We modify their algorithm so that this set of orbit representatives is in fact a fundamental domain, and so that it outputs an explicit projection map. This map  is easy to implement and efficient to compute. In fact, here we will define the \emph{ascending projection} $\pi_\uparrow$, in \cref{sec:rel-proj-maps} we discuss a few variations of this projection map and compute an example. We have listed the outputs of the algorithm for several common examples of groups $G\le S_n$ in \cref{sec:combinatorial-examples}.
	
	\subsubsection{Finding a base}\label{subsec:base}
	
	Let $N=\{1,\dots,n\}$, which we identify with the set of indices for the standard basis for $\R^n$ so $S_n$ acting on $\R^n$ corresponds to the right action of $S_n$ on $N$ by $i\cdot s=s^{-1}(i)$. The first step of the algorithm is to find a \emph{base} for $G\le S_n$.
	
	\begin{definition}
		A \emph{base} for $G\le S_n$ is an ordered subset $B=(b_1,\dots,b_k)$ of $N$ such that
		$
		\bigcap_{i=1}^k\Stab_G(b_i)=\{1\},
		$
		where $\Stab_G(b_i)=\{g\in G\mid b_i\cdot g=b_i\}$ is the \emph{stabiliser} of $b_i$ in $G$. Given a base let $G_0=G$ and for $1\le i\le k$, define $G_i=\Stab_{G_{i-1}}(b_i)=G_{i-1}\cap\Stab_G(b_i)$. 
	\end{definition}
	
	It follows from this definition that $G_k=\{1\}$.  
	Given a base $B$ and the groups $G_i$, we will also define $\Delta_i$ to be the orbit of $b_i$ under the action of $G_{i-1}$.
	
	\begin{example}
		Let $G$ be the subgroup in $S_4$ generated by the elements $(1\;2)$ and $(3\;4)$, where we represent permutations using cycle notation: eg $(1\;2)$ swaps $1$ and $2$ and fixes $3$ and $4$. Then $(b_1=1,b_2=3)$ is a base and we have
		\begin{align*}
			&G_0=\{e,(1\;2),(3\;4),(1\;2)(3\;4)\},\\
			&G_1=\{e,(3\;4)\},\\
			&G_2=\{e\},
		\end{align*}
		and orbits $\Delta_1=\{1,2\}$, and $\Delta_2=\{3,4\}$.
	\end{example}
	
	\subsubsection{Perturbing points in $\R^n$}\label{subsec:perturb}
	
	We now need to define the map $\phi_\uparrow \colon X\to G$ used in the definition of $\pi_\uparrow$. The map $\phi_\uparrow$ will only be uniquely defined on points $x=(x_i)_i\in\R^n$ all of whose entries are distinct. We first perturb $x$ slightly to get a point with this property. Choose a \emph{perturbation vector} $\varepsilon $ which has all distinct entries, for example $\varepsilon =\frac{1}{2n}(1,2,\dots,n).$ Let ${d=\min_{x_i\neq x_j}\{|x_i-x_j|\}}$ (choose $d=1$ if all entries of $x$ are the same) and define $x'=x+d\varepsilon $, which is guaranteed to have all entries distinct.
	
	The entries of $x'$ have the same relative order, ie if ${x'_i\le x'_j}$ then $x_i\le x_j$, and $\phi_\uparrow$ will depend only on this relative ordering of entries. Replace $x'=(x'_1,\dots,x'_n)$ with the point $\hat{x}=(\hat{x}_1,\dots,\hat{x}_n)$ which is defined by ${\hat{x}_i=|\{1\le j\le n\mid x'_j\le x'_i\}|}$. The result will be an ordered list of the integers $1,\dots,n$. Then we define ${\phi_\uparrow(x)=\phi_\uparrow(\hat{x})}$ where $\phi_\uparrow(\hat{x})$ is defined in the next section.
	
	\subsubsection{The ascending projection map}\label{subsec:proj-map}
	
	We will define a sequence of permutations $g_i\in G$ for ${1\le i\le k}$ as follows. Assume $g_1,\dots,g_{i-1}$ have already been found. $G_{i-1}$ acts transitively on $\Delta_i$, choose $j\in\Delta_i$ such that the $j$th entry of $(g_{i-1}\cdots g_1)\cdot\hat{x}$ is minimal among those entries indexed by $\Delta_i$. Choose $g_i\in G_{i-1}$ such that $j\cdot g_i=g_i^{-1}(j)=b_i$. Now define $\phi_\uparrow(\hat{x})\coloneqq g_k\cdots g_1$, note the choice of the $g_i$'s is not unique, but we will show in \cref{subsec:finish-proof} that $\phi(\hat{x})$ \emph{is} uniquely defined.
	
	\Cref{sec:main-proof} is devoted to the proof of the following theorem which says that the map we have defined is a projection onto a fundamental domain.
	
	\begin{theorem}\label{thm:comb-alg}
		Define $\pi_\uparrow \colon \R^n\to \R^n$  by ${\pi_\uparrow(x)=\phi_\uparrow(x)\cdot x}$, and let $\fd$ be the interior of its image. Then $\fd$ is a fundamental domain for $G$ acting on $\R^n$. Given a choice of base $B$ and perturbation vector $\varepsilon $, the projection $\pi_\uparrow$ is uniquely defined.
	\end{theorem}
	
	We can apply this Theorem to the examples we discussed above. 
	
	\setcounter{theorem}{3}
	
	\begin{example}[Continued]
		For the example of $G=S_n$ we can choose as our base ${B=(1,2,\dots,n-1)}$, so that $G_i={\textrm{Perm}(i+1,\dots,n)}$ and $\Delta_i=\{i,\dots,n\}$. Fix $\hat{x}\in X$ and we can follow the algorithm above: $g_1\in G_0=S_n$ is a permutation which moves the smallest entry indexed by $\Delta_1=\{1,\dots,n\}$ to the first entry indexed by $b_1=1$. Next $g_2\in \textrm{Perm}(2,\dots,n)$ moves the next smallest entry, which must be indexed by $\Delta_2=\{2,\dots,n\}$ to be the second entry indexed by $b_2=2$. Repeating this for each $i$ up to $n-1$, $g_i$ moves the $i$th smallest entry of $\hat{x}$ to the $i$th position. The result is that $(g_{n-1}\cdots g_1)\cdot\hat{x}$ has its entries ordered from smallest to largest. 
		
		In a very similar way one can check that applying the algorithm to the case $G=\Z_n$ using the base $B=(1)$ yields the same projection onto a fundamental domain described at the start of \cref{subsec:comb-projections}.
	\end{example}
	
	\setcounter{theorem}{7}
	
	\subsection{Dirichlet projections}\label{subsec:grad-desc}
	
	The second method of computing a projection map is as follows. The action of $G$ leaves the Euclidean inner product $\langle \cdot, \cdot \rangle$ invariant. Choose a point $\hat{x} \in \R^n$ which is only fixed by elements of $G$ which fix the whole of $\R^n$ point-wise. For $x \in \R^n$ define $\phi(x)$ to be an element of $G$ which minimises the function $G\to \R,\; g\mapsto \langle g \cdot x,\hat{x} \rangle$. 
	The function $\pi\colon \R^n\to \R^n,\; x\mapsto \phi(x)\cdot x$ is the projection map onto a fundamental domain which is the image ${\fdc=\pi(\R^n)}$. The fundamental domain constructed in this way is an example of a \textit{Dirichlet fundamental domain}.
	Here, $\phi$ is the minimiser of a linear functional on the discrete space $G$ and can be approximated using a discrete version of gradient descent, see \cref{subsec:implementation-grad-desc}.

	\subsection{Comparing approaches to invariant machine learning} \label{subsec:comparison}
	
	We can compare the various approaches to invariant machine learning which we discussed in \cref{subsec:lit-review} on a theoretical level; in the next section we also compare them experimentally. Data augmentation, like our approach, is a data pre-processing step and so can be applied to any model. However, when using augmentation, the result need not be $G$-invariant, and for large groups it is computationally impractical to augment by a representative sample of elements of the group.
	
	As for intrinsic approaches, group invariant neural networks like \emph{deep sets} \cite{Zaheer2017} are model-specific. The approach in \cite{Yarotsky2021} using polynomial invariants 
	has the drawback that computing the polynomial invariants is not practical for anything other than very simple group actions, and it embeds the training data into a vector space with a much larger dimension than the original data, so there are many more learnable parameters involved making training computationally more expensive. 
	
	
	Projecting onto a fundamental domain combines the benefits of being computationally easy to use, maintaining the original dimension of the data, and being compatible with any machine learning model. In some circumstances the projection fails to be $G$-invariant everywhere (see \cref{remark:boundary}), but for many use-cases this turns out not to be a problem as the data generically lies in the regions where the projection is truly invariant.

	\section{Examples and results}\label{sec:results}
	
	In this section we show how $G$-invariant pre-processing can be applied to examples of classification tasks in group theory, string theory, and image recognition. In each case, the symmetry group acts differently on the input space. We account for this by appropriately choosing different projection maps from the previous section.
	
	\subsection{Cayley tables}\label{subsec:cayley}
	
	Multiplication tables of groups are called \emph{Cayley tables}.
	The following model problem was studied in Section 3.2.3 of \cite{He2019}:
	up to isomorphism, there are $5$ groups with $8$ elements.
	Separate their Cayley tables into two classes and apply random permutations until $20000$ tables in each class exist.
	The problem is then to assign the correct one of two classes to a given table, and this map is invariant under the action of $S_{12} \times S_{12}$ acting on $\R^{12 \times 12}$ by row and column permutations.
	
	Let $\pi_\uparrow \colon  \R^{12 \times 12} \rightarrow \R^{12 \times 12}$ be the ascending projection map from \cref{subsec:comb-projections}, in particular as defined in \cref{subsec:S_n,subsec:tensors}.
	This has an explicit description as follows: given a choice of total order on the group elements, permute the columns so that the first row is ordered smallest to biggest, and then permute the rows so that the first column is ordered smallest to biggest.
	Then, $\pi_\uparrow$ is invariant under the action of $S_{12} \times S_{12}$ and can be efficiently computed for Cayley tables.
	This pre-processing effectively `undoes' the permutations, which makes the machine learning problem trivial.
	Consequently, we achieve nearly perfect accuracy using a linear support vector machine (SVM), see \cref{tab:cayley-accuracy}.
	
	\begin{table}[h]
	\begin{center}
	\begin{tabular}{lc}
	\toprule
	     & Accuracy \\ \midrule
	MLP \cite{He2019}  & $0.501 \pm 0.015$    \\
	$\pi_\uparrow+$SVM & $\mathbf{0.994 \pm 0.008}$   \\ \bottomrule
	\end{tabular}
	\end{center}
	\caption{Accuracy for the task of predicting the group isomorphism type of a Cayley table introduced in Section 3.2.3 of \cite{He2019}.}
	\label{tab:cayley-accuracy}
	\end{table}

	\subsection{CICY}\label{subsec:CICY}	
	
	\begin{table*}[h]
		\begin{center}
			\begin{tabular}{@{}lccccc@{}}
				\toprule
				& 
				Original dataset & 
				Randomly permuted \\ 
				\midrule
				MLP \cite{He2017} & 
				$0.554 \pm 0.015$ 
				& 
				$0.395 \pm 0.029$ 
				\\
				MLP+pre-processing \cite{Bull2019} 
				& 
				$0.858 \pm 0.009$ 
				& 
				$0.417 \pm 0.086$ 
				\\
				Inception \cite{Erbin2021}        & 
				$0.970 \pm 0.009$ 
				& 
				$0.844 \pm 0.117$ 
				\\
				$G$-inv MLP \cite{Hartford2018}        & 
				$0.895 \pm 0.029$ 
				& 
				$0.914 \pm 0.023$ 
				\\
				$\pi_\mathrm{Dir}+$Inception        & 
				$\mathbf{0.975 \pm 0.007}$ 
				& 
				$\mathbf{0.963 \pm 0.016}$ 
				\\
				$\pi_\uparrow+$Inception        & 
				$0.969 \pm 0.009$
				& 
				$0.539 \pm 0.020$ 
				\\ \bottomrule
			\end{tabular}
		\end{center}
		\caption{Prediction accuracies for the task of predicting the second Hodge number of a CICY matrix. Models are compared on the original training task and on randomly permuted input matrices. The last three rows are group invariant models, the first three rows are not group invariant models.}
		\label{table:cicy-results}
	\end{table*}
	
	In \cite{Green1987}, a dataset of complex three-dimensional \emph{complete intersection Calabi-Yau} manifolds (CICYs) and their basic topological invariants is given.
	In \cite{He2017}, a neural network was used to predict (among other tasks) the first Hodge number of a given CICY.
	Here, CICYs are represented by matrices of size up to $12 \times 15$, and the first Hodge number is an integer.
	The same problem was subsequently studied in \cite{Bull2018,Bull2019,Erbin2021}, using more sophisticated machine learning models.
	The problem is invariant under row and column permutations, ie an action of $S_{12} \times S_{15}$ on $\R^{12 \times 15}$, but none of the proposed machine learning models satisfy this invariance.

	We compare two pre-processing maps: 	first, the map $\pi_\mathrm{Dir}\colon  \R^{12 \times 15} \rightarrow \R^{12 \times 15}$ defined in \cref{subsec:grad-desc}, which we computed by performing discrete gradient descent. 
	Second, the map $\pi_\uparrow\colon  \R^{12 \times 15} \rightarrow \R^{12 \times 15}$ defined in the same way as in \cref{subsec:cayley}.
	We found that composing $\pi_\mathrm{Dir}$ with existing neural networks slightly improves performance, but not significantly so.
	We also considered an alternative training task in which random row and column permutations were applied to input matrices before training.
	In this case, our model outperforms models from the literature by a large margin. We also compare our model with the group invariant model from \cite{Hartford2018} in both training tasks, see \cref{table:cicy-results}. 
	
	As our approach is intrinsic it is well suited for problems with a large symmetry group. For all networks but the $G$-invariant multi layer perceptron (MLP) the accuracy decreases on the permuted dataset. This suggests that the rows and columns of the CICY matrices are already systematically ordered in the original dataset.
	The map $\pi_\uparrow$ can be computed efficiently but need not be $G$-invariant on the boundary of the fundamental domain by \cref{remark:boundary}. This is a potential problem since the input data, which consists of integer-valued matrices, is discrete. Indeed, a substantial proportion of the CICY matrices are very sparse and do lie on the boundary, which could be the reason why  $\pi_\uparrow$ performs relatively poorly on the permuted data set. The projection map $\pi_\mathrm{Dir}$ can only be approximated but is fully $G$-invariant which is a crucial advantage on the permuted dataset. 
	 
	\subsection{Classifying rotated handwritten digits}\label{subsec:digits}
	
	As an instructional example, we use a variation of the MNIST dataset of handwritten images from \cite{LeCun1998}, downsized to $8\times 8$ pixels, in which images are acted on by $\Z_4$ by being rotated by multiples of $90^\circ$.
	We use a combinatorial projection map, in particular the ascending averaging projection defined in \cref{sec:rel-proj-maps}, $\pi_{\uparrow\mathrm{av}}\colon  \R^{8 \times 8} \rightarrow \R^{8 \times 8}$.
	The map $\pi_{\uparrow\mathrm{av}}$ rotates each image so that its brightest quadrant is the top-left quadrant.
	We then compare performance of a linear classifier and a shallow neural network; first on their own, then with data augmentation, and finally with the projection map $\pi_{\uparrow\mathrm{av}}$, but without data augmentation, see \cref{tab:mnist-performance}.
	
	For linear classifiers, we observe that data augmentation does not improve accuracy substantially.
	This is due to the small number of parameters for linear classifiers.
	Unsurprisingly, pre-processing with $\pi_{\uparrow\mathrm{av}}$ improves performance, because it is partially successful at rotating digit pictures into a canonical orientation.
	
	For neural networks with more than one layer, data augmentation increases accuracy, because the model now has sufficient parameters to include the information from the additional training data.
	Pre-processing using $\pi_{\uparrow\mathrm{av}}$ yields better accuracy than no pre-processing, but worse accuracy than full data augmentation which adds one new training data point for each possible image rotation.
	If fewer training data points are added during the data augmentation step, the benefit is comparable to applying pre-processing using the map $\pi_{\uparrow\mathrm{av}}$.
	
	This is one example of the fact that data augmentation may be the best pre-processing option if the symmetry group $G$ has few elements and one can augment by the full group.
	If $\left| G \right|$ is very large, this is not possible, and pre-processing using a fundamental domain projection may be better than augmenting with a small, non-representative, subset of $G$.
	
	\begin{table*}[h]
		\begin{center}
			\begin{tabular}{l|cccc}
				\toprule
				& No pre-processing & Augmentation $\times 1.5$ & Augmentation $\times 4$  & $\pi_{\uparrow\mathrm{av}}$    \\ \midrule
				Linear     & $0.631 \pm 0.001$ & $0.632 \pm 0.002$ & $0.633 \pm 0.001$ & $\mathbf{0.756 \pm 0.001}$ \\
				MLP        & $0.941 \pm 0.003$ & $0.948 \pm 0.002$ & $\mathbf{0.961 \pm 0.001}$ & $0.951 \pm 0.002$ \\ \bottomrule
			\end{tabular}
		\end{center}
		\caption{Accuracy for the task of recognising handwritten digits. We compare a linear classifier and a shallow MLP, applying either no pre-processing, or data augmentation, or the pre-processing map $\pi_{\uparrow\mathrm{av}}$ from \cref{subsec:digits}. We use two different degrees of data augmentation: either add each possible rotation of the input picture to the training data (labelled by \emph{Augmentation $\times 4$}) or applying data augmentation until the number of training data points has reached $1.5$ times the original number of training data points (labelled by \emph{Augmentation $\times 1.5$}).}
		\label{tab:mnist-performance}
	\end{table*}

	\section{Unifying intrinsic approaches to equivariant machine learning}\label{sec:unification}
	
	In the previous two sections we approximated $G$-invariant functions by considering them as functions on a fundamental domain.
	In this section we formulate a theorem stating that approximating a $G$-invariant function is equivalent to approximating it on the \emph{quotient space}, and explore how this relates to functions on fundamental domains.
	We then pick up two approaches to $G$-invariant machine learning from the literature alongside our approach, and explain in which sense they can be viewed as machine learning on quotient spaces.
%
	In this section it is convenient to treat the $G$-invariant problem as a special case of the $G$-equivariant problem by observing that an invariant function $X\to Y$ is just an equivariant function where $G$ acts trivially on $Y$. 
	
	\subsection{Universality of quotient spaces}
	
	We begin by defining the quotient space of a group action, and discuss its properties.
	
	\begin{definition}\label{def:quotient-space}
		Let $G$ be a group acting on a set $X$, then the \emph{quotient space} $X/G$ is the set of all $G$-orbits of points in $X$ $G\cdot x=\{g\cdot x\mid g\in G\}$. A quotient space is automatically equipped with a $G$-invariant map 
		\[
		\pi_X\colon X\to X/G\colon x\mapsto G\cdot x.
		\]
		If $X$ is a subset of $\R^n$ and the action is by isometries, $X/G$ inherits a metric from the Euclidean metric (away from some small singular set), and $\pi_X$ is a local isometry.
	\end{definition}
	
	For us the key is that quotient spaces are \emph{universal} with respect to $G$-equivariant maps.
	
	\paragraph{Universal property of quotient spaces} Given two spaces $X$ and $Y$ and a group $G$ acting on both, let $\pi_X\colon X\to X/G$ and $\pi_Y\colon Y\to Y/G$ be the canonical projection maps. Then for any $G$-equivariant map $\al\colon X\to Y$ there is a unique map $\overline{\al}\colon X/G\to Y/G$ making the following diagram commute
	\[
	\begin{tikzcd}
		X \arrow[rr, "\al"] \arrow[d, "\pi_X"] &  &   Y \arrow[d, "\pi_Y"]\\
		X/G \arrow[rr, "\overline{\al}", dashed]  &  & Y/G
	\end{tikzcd}
	\]
	
	If the action of $G$ on $X$ and $Y$ is sufficiently `nice', then the converse holds, ie equivariant maps $X\to Y$ are parametrised by certain maps $X/G\to Y/G$. Here we state this parametrisation in the case of permutation actions on real vector spaces, the main case in this paper, however a version of this holds for a very broad class of group actions. Notice that for any $G$-equivariant function $\alpha$, point stabilisers in $X$ and $Y$ have the property that $\Stab_G(x) \subset \Stab_G(\alpha(x))$, which motivates the following definition.
	
	\begin{definition}\label{def:compatible}
		Let $X$ and $Y$ be (simply connected subsets of) real vector spaces, and suppose $G$ acts on $X$ and $Y$ by permuting coordinates. A continuous map ${\overline{\al}\colon X/G\to Y/G}$ is \emph{compatible} with the $G$ actions if for any $x\in X$, the stabiliser $\Stab_G(x)$ is conjugate in $G$ to a subgroup of $\Stab_G(y)$, where $y$ is a lift under $\pi_Y$ of $\overline{\al}(\pi_X(x))$.
	\end{definition}

	It follows from the observation above that if $\overline{\alpha}$ is the map coming from the universal property, then $\overline{\alpha}$ is automatically compatible with the $G$-action. On the other hand, in the special case that $G$ acts trivially on $Y$ (ie we are in the $G$-invariant case) then every continuous map ${\overline{\al}\colon X/G\to Y/G=Y}$ is compatible. 
	
	\begin{theorem}\label{thm:G-equi-fns}
		Any compatible function $\overline{\alpha}\colon X/G\to Y/G$ lifts to a $G$-equivariant function $\alpha\colon X\to Y$. Suppose $\alpha'$ is another such lift, and assume there is some $x_0\in X$ such that $\alpha'(x_0)=\alpha(x_0)$ where $\Stab_G(\alpha(x_0))$ fixes $Y$ point-wise. Then $\alpha'(x)=\alpha(x)$ for all $x\in X$.
	\end{theorem}
	
	We do not prove this here, but it can be deduced using Theorem 4.1.6 from \cite{Chen2001}. This gives a converse to the universal property, and shows that up to acting globally on $Y$ by a fixed element of $G$ there is a one-to-one correspondence between $G$ equivariant maps of vector spaces, and compatible maps of their quotients. 
	
	Now suppose $\al\colon X\to Y$ is a $G$-equivariant function we want to approximate using a supervised machine learning algorithm. Using an intrinsic approach, as discussed in  \cref{subsec:lit-review}, means approximating $\al$ by a function $\be$ which is \emph{a priori} $G$-equivariant. By the theorem above, this is equivalent to approximating $\overline{\al}$ by a compatible function $\overline{\be}$. Below we will discuss how different intrinsic approaches to the equivariant machine learning problem fit into this framework.
	
	\subsection{Equivariant layers in neural networks}
	\label{subsec-equiv-layers}
	A $G$-equivariant neural network (see for example \cite{Maron2020}) consists of a series of $G$-equivariant linear maps $\lambda_i$ separated by some non-linear activation function $\sigma$, yielding $\be=\lambda_k\circ\sigma\circ\cdots\circ\sigma\circ\lambda_1$. Restrictions are placed on the learnable parameters of each $\lambda_i$ to ensure they are $G$-equivariant. For example, if $\lambda_i\colon \R^n\to\R^n$ is equivariant with respect to $S_n$ acting on each copy of $\R^n$ by permuting coordinates, then it was shown in Lemma 3 of \cite{Zaheer2017}, that it must have the form 
	\[
	\lambda_i(x)=(a\mathbb{I}+b\mathds{1}^T\mathds{1})x,
	\] 
	where $a,b\in\R$ are learnable parameters, $\mathbb{I}$ is the identity matrix, and ${\mathds{1}=(1,1,\dots,1)}$.
	The main task is to describe the space of all $G$-equivariant linear maps $\lambda\colon \R^{n_1}\to\R^{n_2}$ which could map between layers in the neural network. There are two powerful approaches to tackle this problem. One is based on decomposing the representation of $G$ on $\R^{n_1}$ and $\R^{n_2}$ into irreducible components and applying Schur's lemma. The other approach is combinatorial and involves putting a $G$-invariant simplicial complex structure on $\R^{n_i}$ and applying the compatibility criterion to the cells in the simplicial structure induced on the quotient space, see \cref{subsec:comp-lin-equiv}.
	
	\subsection{Equivariant maps from polynomial invariants}\label{subsec:poly-inv}
	
	We discuss briefly the approach proposed by Yarotsky in Section 2 of \cite{Yarotsky2021} based on the theory of polynomial invariants of $G$. 
	
	\begin{definition}
		For a group $G$ acting on $\R^n$, a polynomial $p(x)\in \R[x_1,\dots,x_n]$ is called a \emph{polynomial invariant} for $G$ if, viewed as a function $p\colon \R^n\to \R$, it satisfies \cref{equation:invariance-property}. Similarly, if $G$ also acts on $\R^m$, then  $q(x)\colon \R^n\to\R^m$ is a \emph{polynomial equivariant} if $q(g\cdot x)=g\cdot q(x)$ for all $g\in G$ and $x\in\R^n$.
	\end{definition}
	 
	The following, proved in \cite{Noether1915} in the invariant case, was generalised to compact Lie groups in Theorem 8.14.A of \cite{Weyl1966}. The equivariant case is proved in Section 4 of \cite{Worfolk1994}.
	If $G$ is a finite group acting on $\R^n$ and $\R^m$, then there are finite sets of polynomial invariants $\{p_i(x)\}_{i=1}^k$, and polynomial equivariants $\{q_j(x)\}_{j=1}^l$ such that any polynomial equivariant $q(x)$ can be expressed as $q(x)=\sum_{j=1}^lq_j(x)r_j(p_1(x),\dots,p_k(x))$ for some $r_j(x)\in \R[x_1,\dots,x_k]$. In the invariant case we can take $l=1$, and $q_1(x)=1$.
	
	Yarotsky shows in Proposition 2.4 in \cite{Yarotsky2021} that any continuous $G$-equivariant function $\al\colon \R^n\to\R^m$ can be approximated on a compact set by a $G$-equivariant neural network of the form
	\begin{equation}\label{eq:G-equi-NN}
	\be(x)=\sum_{j=1}^lq_j(x)\sum_{h=1}^da_{jh}\,\sigma\!\left(\sum_{i=1}^kb_{jhi}p_i(x)+c_{jh}\right)
	\end{equation}
	for some $a_{jh},b_{jhi},c_{jh}\in\R$, where $d\in \mbN$, and $\sigma$ is a continuous non-polynomial activation function. 
	
	Notice that all the learnable parameters are contained in the inner two sums, which also constitute the neural network in the $G$-invariant case $\sum_{h=1}^da_{h}\,\sigma\!\left(\sum_{i=1}^kb_{hi}p_i(x)+c_{h}\right)$, see Proposition 2.3 in \cite{Yarotsky2021}. The relationship between this method and quotient spaces is shown by the following result \cite{Schwarz1975}.
	
	\begin{theorem}\label{thm:Schwarz}
		The map $p(x)\coloneqq(p_1(x),\dots,p_k(x))$ factors through $\R^n/G$, and induces a smooth embedding of $\R^n/G$ into $\R^k$.
	\end{theorem}
	
	We can reinterpret the invariant version of \cref{eq:G-equi-NN} as a fully-connected neural network $\overline{\be}$.
	We then train this network on the data which has been projected to the quotient space by $p$, $D^{p}_{\mathrm{train}}=\{(p(x),y)\mid (x,y)\in D_{{\mathrm{train}}}\}$. In this way, $\overline{\be}$ learns the map $\overline{\al}$ produced by the universal property of quotient spaces.
	
	\subsection{Our approaches}
	
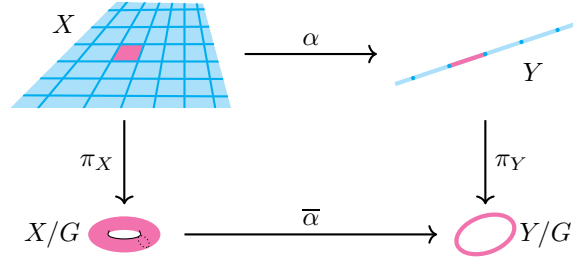
\begin{figure}[h]
		\centering
		\begin{tikzpicture}[scale=.8]
			\tikzset{
				partial ellipse/.style args={#1:#2:#3}{
					insert path={+ (#1:#3) arc (#1:#2:#3)}
				}
			}
			
			\begin{scope}
				\node at (-2,-1.5) {$X$};
				\clip(285:3) -- (225:4.1) -- (225:1.6) -- (285:1.2) -- cycle;
				\fill[cyan!30] (285:3) -- (225:4.1) -- (225:1.6) -- (285:1.2) -- cycle;
				\fill[magenta!70] (-1.05,-1.85) -- (-1.23,-2.13) -- (-.78,-2.13) -- (-.68,-1.85) -- cycle;
				\foreach \i in  {0,...,5}{
					\draw[thick,cyan] (0,0) -- (280-10*\i:4);
					\draw[thick,cyan] (225:1.8+.4*\i) -- (285:1.3+.3*\i);
				}
			\end{scope}
			
			\begin{scope}[yshift=-5cm,xshift=-1cm]
				\node at (-1.2,0) {$X/G$};
				\fill[magenta!70] (0,0) ellipse (.6 and .3);
				\fill[white] (0,0) ellipse (.27 and .075);
				\draw (0,0) [partial ellipse=180:360:.27 and .075];
				\draw (-.17,0) [partial ellipse=140:182:.1 and .1];
				\draw (.17,0) [partial ellipse=-2:40:.1 and .1];
				\draw[,densely dotted,rotate=-50] (.3,.17) ellipse (.115 and .07);
			\end{scope}
			
			\begin{scope}[yshift=-2.5cm,xshift=3.5cm]
				\node at (2.3,.2) {$Y$};
				\draw[ultra thick, cyan!30] (0,0) -- (3,1);
				\draw[ultra thick, magenta!70] (0.9,.3) -- (1.5,.5);
				\foreach \i in {0,...,4}{
					\fill[cyan] (.3+.6*\i,.1+.2*\i) circle (1pt);	
				}
			\end{scope}
			
			\begin{scope}[yshift=-5cm,xshift=5cm]
				\node at (1,0) {$Y/G$};
				\draw[ultra thick, magenta!70, rotate=20] (0,0) ellipse (.5 and .3);
			\end{scope}
			
			\draw[thick,->] (-1,-3.1) -- (-1,-4.5);
			\node at (-1,-3.8) [anchor=east]{$\pi_X$};
			
			\draw[thick,->] (5,-3.1) -- (5,-4.5);
			\node at (5,-3.8) [anchor=west]{$\pi_Y$};
			
			\draw[thick,->] (1,-2) -- (3.2,-2);
			\node at (2.1,-2) [anchor=south]{$\alpha$};
			
			\draw[thick,->] (0,-5) -- (4.2,-5);
			\node at (2.1,-5) [anchor=south]{$\overline{\alpha}$};
		\end{tikzpicture}
		\caption{A fundamental domain can be thought of as a chart for the quotient space.}
		\label{figure:quotient-space}
	\end{figure}

	Our approach of projecting onto a fundamental domain (whether by a combinatorial projection or a Dirichlet projection) fit very naturally in this general framework. Like Yarotsky's approach, we want to try to approximate the function $\overline{\al}$ rather than approximating $\al$. Instead of working directly with the quotient spaces, one can think of the map from the fundamental domain to the quotient space $\restrict{\pi_X}{\fd_X}\colon \fd_X\to X/G$ as a \emph{chart} in the sense of differential geometry, and so $\fd_X$ locally parametrises $X/G$, see \cref{figure:quotient-space}. 
	
	In the invariant case, we can approximate $\al=\overline{\al}\circ\restrict{\pi_X}{\fd_X}$ by approximating $\overline{\al}$. In the equivariant case, we can also view $\restrict{\pi_Y}{\fd_Y}\colon \fd_Y\to Y/G$ as a chart, and because $\restrict{\pi_Y}{\fd_Y}$ is a bijection onto its image we can apply its inverse and approximate $\al=(\restrict{\pi_Y}{\fd_Y})^{-1}\circ\overline{\al}\circ\restrict{\pi_X}{\fd_X}$ by approximating $\overline{\al}$. Note that $\restrict{\pi_X}{\fd_X}$ and $\restrict{\pi_Y}{\fd_Y}$ are not surjective in general, and there is no canonical way to extend their domain to make them so. The fix for this is to perturb points to lie in the preimage of $\fd_X$ as discussed in \cref{subsec:perturb}.
	
	\begin{remark}
		We do not work directly with the quotient spaces in our approach because the quotient space of a vector space is not itself a vector space. It needs to be embedded first in another vector space before being used as the input for a neural network, say. Finding quotient space embeddings is an extremely difficult problem. The approach in \cite{Yarotsky2021} is to use polynomial invariants, which can be found using an algorithm. The problem, in addition to being computationally infeasible in practice, is that they significantly distort the training data leading to low accuracy. To avoid this one must find an \emph{isometric} embedding which does not distort the data. However, this is even more difficult, and likewise significantly increases the ambient dimension of the training data. 
	\end{remark}

	\section{Conclusion}\label{sec:conclusion}
	The $G$-invariant pre-processing step proposed in this paper has a clear mathematical motivation. It respects the geometry of the input space and as seen in \cref{sec:unification} it naturally fits into a larger framework of $G$-equivariant machine learning.
	There are also many practical advantages of our approach: it can be applied to any machine learning architecture, it preserves the dimension of the input space and in most cases it guarantees perfect $G$-invariance. Furthermore, the computation cost is generally low even if $|G|$ is very large.
	For the image recognition task, $|G|=4$, so many of these advantages are only relevant for networks with a small number of neurons. 
	
	For Cayley tables however, $|G|=12!^2\approx 2 \times 10^{17}$ is very large and the $G$-invariance of our approach produces nearly perfect accuracy. The symmetry group is even larger for CICY matrices even though both the action of the symmetry group as well as the classification task itself are more complicated. Our approach significantly improves the most accurate architecture known so far for this task.

\bibliography{bibliography/bibliography}
\bibliographystyle{icml2022}

\newpage
\appendix
\onecolumn

	\section{Other combinatorial projection maps}\label{sec:rel-proj-maps}
	
	There are three natural variations of the combinatorial projection map $\pi_\uparrow$ we defined in \cref{subsec:proj-map} which may be more suited to specific applications. We called that projection an \emph{ascending projection}. The variations are a \emph{descending projection} $\pi_\downarrow$, and ascending and descending \emph{averaging projections} $\pi_{\uparrow\mathrm{av}}$ and $\pi_{\downarrow\mathrm{av}}$. These projections each have their own version of \cref{thm:comb-alg} whose proof is essentially identical.
	
	The descending projection is defined via $\phi_\downarrow$, which differs from $\phi_\uparrow$ only when we define $g_i$. In this case $G_{i-1}$ acts transitively on $\Delta_i$, and we choose $j\in\Delta_i$ such that the $j$th entry of $(g_{i-1}\cdots g_1)\cdot\hat{x}$ is \emph{maximal} among those entries indexed by $\Delta_i$. Choose $g_i\in G_{i-1}$ such that $j\cdot g_i=g_i^{-1}(j)=b_i$. If the input data for the machine learning algorithm consisted of vectors containing non-negative entries including many zeros, the descending projection in some sense \emph{prioritises} the non-zero entries, so may yield different results. 
	
	For the averaging projections, assume that $G=H_1\times H_2$ is a direct product of groups $H_j\le S_{n_j}$ which acts the space of $n_1\times n_2$ matrices, $\R^{n_1}\otimes\R^{n_2}$, by letting $H_1$ permute rows and $H_2$ permute columns. In this case, identify $N$ with the set of pairs $\{(l,m)\mid1\le l\le n_1,\;1\le m\le n_2\}$. Define a transformation $\mu\colon \R^{n_1}\otimes\R^{n_2}\to \R^{n_1}\otimes\R^{n_2}$ by
	\[
	\mu\colon (x_{lm})_{lm}\mapsto\left(\frac{1}{n_1}(x_{1m}+x_{2m}+\cdots+x_{n_1m})+\frac{1}{n_2}(x_{l1}+x_{l2}+\cdots+x_{ln_2})\right)_{lm}.
	\]
	Notice this is a $G$-equivariant linear map which replaces each entry of $(x_{lm})_{lm}$ by the sum of the averages of the entries in its row and column. Now for any $x\in\R^{n_1}\otimes\R^{n_2}$ we define
	\[
	\phi_{\uparrow\mathrm{av}}(x)=\phi_\uparrow\left(\widehat{\mu(x)}\right)\quad\textrm{and}\quad\phi_{\downarrow\mathrm{av}}(x)=\phi_\downarrow\left(\widehat{\mu(x)}\right).
	\]
	These definitions generalise in the obvious way to the case $G=\prod_{j=1}^rH_j$ acting on $\bigotimes_{j=1}^r\R^{n_j}$ component-wise, where $H_j\le S_{n_j}$. One might wish to use an averaging projection if, for example, one of the $H_j$'s is trivial, in which case a non-averaging projection ignores most of the entries, since they will not be in any of the orbits $\Delta_i$. This is the case in the application discussed in \cref{subsec:digits}.
	
	\begin{example}
		Let $G=\Z_3\times S_3\le S_3\times S_3$ act on $\R^{3}\otimes\R^{3}$, thought of as the set of $3\times 3$ matrices,  by cyclically permuting the rows and freely permuting the columns. We let $N=\{(l,m)\mid 1\le l\le 3,\;1\le m\le 3\}$ and construct a base. Let $b_1=(1,1)$ whose stabiliser is $G_1=\{1\}\times\Sym(\{2,3\})$, and the orbit of $b_1$ under $G_0=G$ is $\Delta_1=N$. Now $(2,1)$ and $(3,1)$ are both fixed by $G_1$ and so should not be the next element of the base. Choose $b_2=(1,2)$. Then $G_2=\{1\}\times\{1\}$ and the orbit of $b_2$ under $G_1$ is $\Delta_2=\{(1,2),(1,3)\}$. Since $G_2\cong\{1\}$ we are done and $B=((1,1),(1,2))$.
		
		Let $\hat{x}=\left(\hat{x}_{lm}\right)_{lm}$ be a $3\times 3$ matrix whose set of entries is $\{1,\dots,9\}$, we want to compute $\phi_\uparrow(\hat{x})$. Let $(p_1,q_1)\in \Delta_1=N$ be the pair such that $\hat{x}_{p_1q_1}=1$ is the minimal entry in $\hat{x}$. Then we can choose $g_1=(s_1,(1\;q_1))\in\Z_3\times S_3$ where 
		\[
		s_1=\begin{cases}
			(1)&p_1=1\\
			(1\;2\;3)&p_1=2\\
			(1\;3\;2)&p_1=3
		\end{cases}\;\in\Z_3
		\]
		
		Now let $g_1\cdot\hat{x}=\left(\hat{x}'_{lm}\right)_{lm}$, and let $(1,q_2)\in\Delta_2=\{(1,2),(1,3)\}$ minimise $\hat{x}'_{1q_2}$. Define $g_2=((1),(2\;q_2))\in G_1$ and $\phi_\uparrow(\hat{x})=g_2g_1$. 
		
		Combinatorially we can describe the projection $\pi_\uparrow$ as follows: transport the smallest entry of $\hat{x}$ to the top left corner by cyclically permuting rows and freely permuting columns, and then order columns 2 and 3 so that the entries in the first row increase. 
		
		As an example, consider the matrix $x$ and perturbation matrix $\varepsilon $
		\[
		x=\begin{pmatrix}
			5&3&3\\
			4&0&0\\
			3&5&1
		\end{pmatrix}\quad\quad\varepsilon =\frac{1}{18}\begin{pmatrix}
			1&2&3\\
			4&5&6\\
			7&8&9
		\end{pmatrix}.
		\]
		Then we compute
		\[
		x'=x+\varepsilon =\frac{1}{18}\begin{pmatrix}
			91&56&57\\
			76&5&6\\
			61&98&27
		\end{pmatrix}\Longrightarrow\hat{x}=\begin{pmatrix}
			8&4&5\\
			7&1&2\\
			6&9&3
		\end{pmatrix}.
		\]
		We can now apply $\pi_\uparrow(x)=\phi_\uparrow(\hat{x})\cdot x$ in the two step process described above:
		\[
		\hat{x}=\begin{pmatrix}
			8&4&5\\
			7&1&2\\
			6&9&3
		\end{pmatrix}\overset{g_1}{\longmapsto}\begin{pmatrix}
			1&7&2\\
			9&6&3\\
			4&8&5
		\end{pmatrix}\overset{g_2}{\longmapsto}\begin{pmatrix}
			1&2&7\\
			9&3&6\\
			4&5&8
		\end{pmatrix}=\pi_\uparrow(\hat{x})
		\]
		\[
		x=\begin{pmatrix}
			5&3&3\\
			4&0&0\\
			3&5&1
		\end{pmatrix}\overset{g_1}{\longmapsto}\begin{pmatrix}
			0&4&0\\
			5&3&1\\
			3&5&3
		\end{pmatrix}\overset{g_2}{\longmapsto}\begin{pmatrix}
			0&0&4\\
			5&1&3\\
			3&3&5
		\end{pmatrix}=\pi_\uparrow(x).
		\]
		Similarly
		\[
		\pi_\downarrow(x)=\begin{pmatrix}
			5&3&1\\
			3&5&3\\
			0&4&0
		\end{pmatrix}.
		\]
		We can also compute the averaging versions of these projections
		\[
		\mu(x)=\frac{1}{3}\begin{pmatrix}
			23&19&15\\
			16&12&8\\
			21&17&13
		\end{pmatrix}\Longrightarrow\pi_{\uparrow\mathrm{av}}(x)=\begin{pmatrix}
			0&0&4\\
			1&5&3\\
			3&3&5
		\end{pmatrix},\;\textrm{and}\;\pi_{\downarrow\mathrm{av}}(x)=\begin{pmatrix}
			5&3&3\\
			4&0&0\\
			3&5&1
		\end{pmatrix}.
		\]
	\end{example}

	\section{Examples of combinatorial projection maps}\label{sec:combinatorial-examples}
	
	In this section we list combinatorial projection maps for several common examples of groups $G\le S_n$. Notice that in each of the four examples of concrete groups below, implementation via a suitable sorting function circumvents the need to perturb inputs initially. 
	
	\subsection{The symmetric group}\label{subsec:S_n}
	If $G= S_n$, let $N=\{1,\dots,n\}$ and we can choose the base $B=(1,2,\dots,n-1)$. The ascending projection $\pi_\uparrow(x)$ permutes the entries so that they increase from left-to-right, and the descending projection $\pi_\downarrow(x)$ permutes the entries so that they decrease. 
	
	\subsection{The alternating group}
	
	If $G=A_n<S_n$ is the group of even permutations, we can choose $B=(1,2,\dots,n-2)$ and the ascending (resp.\ descending) projection permutes the entries of $x$ so that the first $n-2$ entries increase (resp.\ decrease) from left-to-right, and the last two entries are greater than or equal to all the other entries. If $x$ contains repeated entries then the last to entries can also be ordered to be increasing (resp. decreasing); otherwise their relative order depends on whether the permutation $s_{\hat{x}}$ which maps $i\mapsto\hat{x}_i$ for $1\le i\le n$, is an even or odd permutation (see \cref{subsec:perturb} for the definition of $\hat{x}_i$). 
	
	\subsection{The cyclic group}
	
	If $G=\Z_n\le S_n$ is the cyclic group generated by the permutation $(1\;2\;\cdots\;n)$, we can choose the base $B=(1)$. The ascending (resp.\ descending) projection cyclically permutes the entries of $x$ so that the first entry is less (resp.\ greater) than or equal to all other entries of $x$.
	
	\subsection{The dihedral group}\label{subsec:dihedral}
	
	If $G=D_n\le S_n$ is the dihedral group generated by 
	\[
	s_1=(1\;2\;\cdots\;n)\quad \textrm{and} \quad s_2=(2\;n)(3\;(n-1))(4\;(n-2))\cdots,
	\]
	we can choose base $B=(1,2)$. The ascending (resp.\ descending) projection cyclically permutes the entries of $x$ via $s_1$ so that the first entry is less (resp.\ greater) than or equal to all other entries of $x$, and then if the final entry is less (resp.\ greater) than the second entry, it applies the permutation $s_2$. 
	
	\subsection{Products of groups acting on products of spaces}
	
	Suppose $G=\prod_{j=1}^rH_j$ where $H_j\le S_{n_j}$ acts on $\bigoplus_{j=1}^r\R^{n_j}$ by each $H_j$ acting by permutations on the corresponding space $\R^{n_j}$ and trivially everywhere else. Let $B_j=\left(b_j^{(1)},\dots,b_j^{(k_j)}\right)\subset\{1,\dots,n_j\}=N_j$ be a base for $H_j$ acting on $\R^{n_j}$, then 
	\[
	B=\left (b_1^{(1)},\dots,b_1^{(k_1)},b_2^{(1)},\dots,b_2^{(k_2)},\dots,b_r^{(1)},\dots,b_r^{(k_r)}\right )
	\]
	is a base for $G$. Let $\pi_{j\uparrow}\colon \R^{n_j}\to \R^{n_j}$ be the ascending projection corresponding to $B_j$. Then define $\pi_\uparrow=\bigoplus_{j=1}^r\pi_{j\uparrow}$, to be the projection which equals $\pi_{j\uparrow}$ when restricted to $\R^{n_j}$. Similarly $\pi_\downarrow=\bigoplus_{j=1}^r\pi_{j\downarrow}$.
	
	\subsection{Products of groups acting on tensors of spaces}\label{subsec:tensors}
	
	
	Suppose $G=\prod_{j=1}^rH_j$ where $H_j\le S_{n_j}$ acts on $\bigotimes_{j=1}^r\R^{n_j}$ by each $H_j$ acting by permutations on the $j$th component of $\bigotimes_{j=1}^r\R^{n_j}$, and trivially on the other components. For each $1\le j\le r$ let $B_j=\left (b_j^{(1)},\dots,b_j^{(k_j)}\right)\subset\{1,\dots,n_j\}=N_j$ be a base for $H_j$ acting on $\R^{n_j}$, and furthermore (for convenience) assume that $b_j^{(1)}=1$. Then choose $B\subset\prod_{j=1}^rN_j\eqqcolon N$ to be
	\begin{align*}
	B=\Bigl((1,\dots,1),\;&\left (b_1^{(2)},1,\dots,1\right ),\dots,\left (b_1^{(k_1)},1,\dots,1\right ),\\
	&\hphantom{(b_1^{(2)},1,}\vdots\hphantom{\dots,1),\dots,(b_1^{(k_1)},1}\vdots\\
	&\left (1,\dots,1,b_r^{(2)}\right ),\dots,\left (1,\dots,1,b_r^{(k_r)}\right )\Bigr),
	\end{align*}
	where a $1$ in the $j$th position of an element of $B$ should be thought of as $b_j^{(1)}$. Suppose $x=(x_{l_1\cdots l_r})_{l_1\cdots l_r}\in \bigotimes_{j=1}^r\R^{n_j}$, and let $\hat{x}$ be defined as in \cref{subsec:perturb}. Choose $(m_1,\dots,m_r)\in N$ to be the index in the $G$-orbit of $\mathds{1}=(1,\dots,1)$ with minimal entry in $\hat{x}$. For $1\le j\le r$ define $\hat{x}_j\coloneqq (\hat{x}_{m_1\cdots l_j\cdots m_r})_{1\le l_j\le n_j}\in \R^{n_j}$, which is the restriction of $\hat{x}$ to the $\R^{n_j}$-vector containing the entry $\hat{x}_{m_1\cdots m_j\cdots m_r}$. Then define
	\[
	\phi_\uparrow\colon \textstyle\bigotimes_{j=1}^r\R^{n_j}\to G\colon x\mapsto (\phi_{1\uparrow}(\hat{x}_1),\dots,\phi_{r\uparrow}(\hat{x}_r)),
	\]
	where $\phi_{j\uparrow}\colon \R^{n_j}\to H_j$ is the function defined for $H_j$ acting on $\R^{n_j}$, and similarly define $\phi_\downarrow(x)$. Then as before, $\pi_\uparrow(x)\coloneqq \phi_\uparrow(x)\cdot x$ and $\pi_\downarrow(x)\coloneqq \phi_\downarrow(x)\cdot x$.

	\section{Proof of \cref{thm:comb-alg}}\label{sec:main-proof}
	
	The idea of the proof is as follows. In \cref{subsec:actns-on-R-S} we shall outline an equivalence between subgroups of $S_n$ acting on $\R^n$ by permuting coordinates, and them acting on $S_n$ by multiplication. This will provide a dictionary between certain combinatorially defined fundamental domains and sets of coset representatives satisfying simple algebraic properties, \cref{prop:G-fund-dom}. We will then outline the work from \cite{Dixon1988} in \cref{subsec:coset-reps} which gives an algorithm to find a set of coset representatives for an arbitrary subgroup of $S_n$. The main work is then to show this algorithm, with modifications, can produce a set of coset representatives with the desired algebraic properties so that it corresponds to a fundamental domain, culminating in \cref{cor:fund-dom}. Finally we will show in \cref{prop:pi-img} that the algorithm outlined in \cref{subsec:comb-projections} indeed produces a projection onto this fundamental domain.
	
	\subsection{Actions on $\R^n$ and $S_n$}\label{subsec:actns-on-R-S}
	
	Recall we have the group $S_n$ acting on $\R^n$ on the left by $s\cdot (x_i)_i=\left(x_{s^{-1}(i)}\right)_i$. We also have the normal action of $S_n$ on itself on the left by group multiplication: $s$ acts on $t$ by $s\cdot t=st$ for any $s,t\in S_n$. Here we shall show that in some sense these actions are equivalent. This correspondence is known, at least to experts, so we will only outline the essential points.
	
	Let $x\in\R^n$ be a point, all of whose entries are distinct, and notice the set of such points is open and dense in $\R^n$. As we did in \cref{subsec:perturb}, define a function which changes the $i$th entry $x_i$ of $x$ to the integer $|\{1\le j\le n\mid x_j\le x_i\}|$. The result will be a list of the integers $1,\dots,n$ in some order, and we denote the set of these points $C$. We can think of $C$ as a discrete subset of $\R^n$, and the left action of $S_n$ on $\R^n$ restricts to a left action on $C$. Notice also that this map $\R^n_\textrm{dist}\coloneqq\{x\in\R^n\mid\textnormal{all entries are distinct}\}\to C$ is continuous. In other words the set of connected components of $\R^n_\textrm{dist}$ is in one-to-one correspondence with $C$, and indeed each component contains a point in $C$, its \emph{representative point}. We call these connected components \emph{chambers}, and given $c\in C$ we will write $[c]\subset\R^n$ for the corresponding chamber. The following is easy to check.
	
	\begin{lemma}\label{lemma:Sn-fund-dom}
		Each chamber is a fundamental domain for the action of $S_n$ on $\R^n$.
	\end{lemma}
	
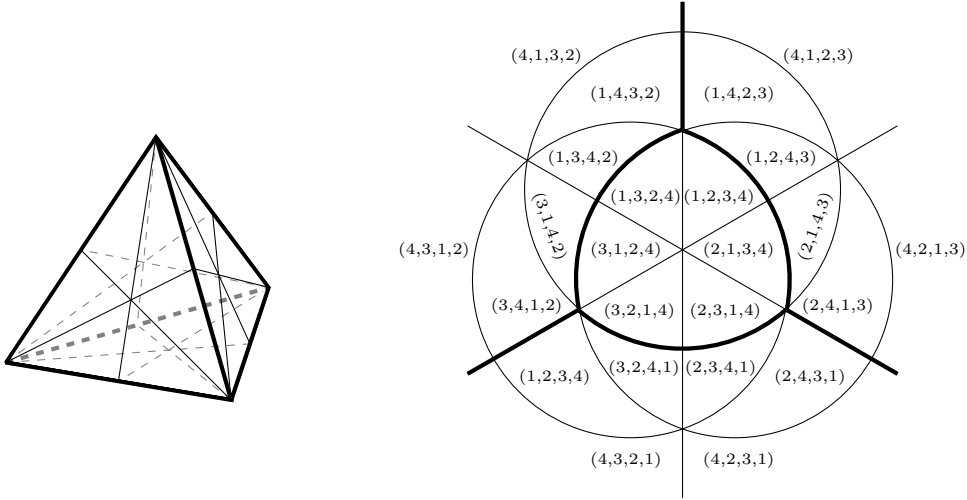
\begin{figure}
		\centering
		\begin{tikzpicture}
			\newcommand\barytriangle[7]{
				\draw (#1,#2) -- (#3,#4) -- (#5,#6) -- cycle;
				\draw (#1,#2) -- (#3/2+#5/2,#4/2+#6/2);
				\draw (#3,#4) -- (#1/2+#5/2,#2/2+#6/2);
				\draw (#5,#6) -- (#1/2+#3/2,#2/2+#4/2);
				\fill (#1,#2) circle (#7pt);
				\fill (#3,#4) circle (#7pt);
				\fill (#5,#6) circle (#7pt);
				\fill (#1/2+#3/2,#2/2+#4/2) circle (#7pt);
				\fill (#5/2+#3/2,#6/2+#4/2) circle (#7pt);
				\fill (#1/2+#5/2,#2/2+#6/2) circle (#7pt);
				\fill (#1/3+#3/3+#5/3,#2/3+#4/3+#6/3) circle (#7pt);
			}
			
			\begin{scope}
				
				\clip (-0.04,-0.02) -- (2.0,3.05) -- (3.53,1) -- (3.03,-0.53) -- cycle;
				
				\begin{scope}[gray, dashed]
					\barytriangle{0}{0}{2}{3}{3.5}{1}{0}
					\barytriangle{3.5}{1}{0}{0}{3}{-0.5}{0}
				\end{scope}
				
				\draw[ultra thick, white] (0,0) -- (3.5,1);
				\draw[ultra thick, gray, dashed] (0,0) -- (3.5,1);
				
				\draw[ultra thick] (0,0) -- (2,3) -- (3.5,1) -- (3,-0.5) -- (0,0) -- (3,-0.5) -- (2,3);
				\barytriangle{0}{0}{2}{3}{3}{-0.5}{0}
				\barytriangle{3.5}{1}{2}{3}{3}{-0.5}{0}
			\end{scope}
			
			\begin{scope}[xshift=9cm,yshift=1.5cm]
				\foreach \i in {0,1,2}{
					\draw (90+\i*120:0.8) circle (2.1);
					\draw (90+\i*120:3.3) -- (-90+\i*120:3.3);
					
					\draw[ultra thick] (90+\i*120:1.6) arc (109+\i*120:191+\i*120:2.1);
					\draw[ultra thick] (90+\i*120:1.6) -- (90+\i*120:3.3);
				}
				
				\node at (0:.75) []{\tiny$(2,\!1,\!3,\!4)$};
				\node at (55:.85) []{\tiny$(1,\!2,\!3,\!4)$};
				\node at (125:.85) []{\tiny$(1,\!3,\!2,\!4)$};
				\node at (180:.75) []{\tiny$(3,\!1,\!2,\!4)$};
				\node at (235:1.0) []{\tiny$(3,\!2,\!1,\!4)$};
				\node at (305:1.0) []{\tiny$(2,\!3,\!1,\!4)$};
				
				\node at (10:1.8) [rotate=70]{\tiny$(2,\!1,\!4,\!3)$};
				\node at (43:1.8) []{\tiny$(1,\!2,\!4,\!3)$};
				\node at (137:1.8) []{\tiny$(1,\!3,\!4,\!2)$};
				\node at (170:1.8) [rotate=-70]{\tiny$(3,\!1,\!4,\!2)$};
				\node at (252:1.65) []{\tiny$(3,\!2,\!4,\!1)$};
				\node at (288:1.65) []{\tiny$(2,\!3,\!4,\!1)$};
				
				\node at (-20:2.2) []{\tiny$(2,\!4,\!1,\!3)$};
				\node at (70:2.2) []{\tiny$(1,\!4,\!2,\!3)$};
				\node at (110:2.2) []{\tiny$(1,\!4,\!3,\!2)$};
				\node at (200:2.2) []{\tiny$(3,\!4,\!1,\!2)$};
				\node at (225:2.4) []{\tiny$(1,\!2,\!3,\!4)$};
				\node at (315:2.4) []{\tiny$(2,\!4,\!3,\!1)$};
				
				\node at (0:3.3) []{\tiny$(4,\!2,\!1,\!3)$};
				\node at (55:3.15) []{\tiny$(4,\!1,\!2,\!3)$};
				\node at (125:3.15) []{\tiny$(4,\!1,\!3,\!2)$};
				\node at (180:3.3) []{\tiny$(4,\!3,\!1,\!2)$};
				\node at (255:2.9) []{\tiny$(4,\!3,\!2,\!1)$};
				\node at (285:2.9) []{\tiny$(4,\!2,\!3,\!1)$};
			\end{scope}
		\end{tikzpicture}
		\caption{On the left is boundary of a $3$-simplex, each small triangle corresponds to the intersection of this with a chamber. On the right the picture has been stereographically projected to the plane for the purposes of illustration, and each chamber is labelled by the representative element of $C$.}
		\label{fig:chambers}
	\end{figure}

	The action of $S_n$ on $\R^n$ preserves an $(n-1)$-simplex in the orthogonal complement of the vector $(1,\dots,1)$. In \cref{fig:chambers} we show the 3-simplex preserved by $S_4$, and use it to visualise the $24=|S_4|$ chambers in this case.
	
	On the other hand, we can view each element of $C$ as a permutation in $S_n$ written in \emph{in-line notation}. This means if $c=(c_i)_i$, as a permutation it sends $i$ to $c_i$ for each $i\in\{1,\dots,n\}$. Thus $S_n$ is in one-to-one correspondence with $C$. In fact, it is better in our situation to modify this correspondence by inverting elements of $S_n$ via the map $\rho\colon S_n\to C\colon s\mapsto (s^{-1}(i))_i$. 	
	The equivalence of the left action of $S_n$ on $\R^n$ and the left action on itself comes in the following form. Let $s,t\in S_n$, and consider the action of $s$ on $\rho(t)$:
	\[
	s\cdot\rho(t)=s\cdot (t^{-1}(i))_i =(t^{-1}(s^{-1}(i)))_i =((st)^{-1}(i))_i =\rho(st) =\rho(s\cdot t).
	\]
	
	Given any subgroup $G\le S_n$, the map $\rho$ defines an equivalence between $G$ acting on $\R^n$, which restricts to an action of $G$ on $C$, and $G$ acting on $S_n$ by left multiplication. We can use this equivalence to convert a set of right coset representatives for $G$ in $S_n$ into a complete set of orbit representatives (defined in \cref{subsec:fun-dom}) for $G$ acting on $\R^n$.
	
	\begin{proposition}\label{prop:G-orbit-reps}
		Let $R$ be a set of right coset representatives for $G\le S_n$, then $\fdc=\bigcup_{r\in R}\overline{[\rho(r)]}$ is a complete set of orbit representatives for $G$ acting on $\R^n$, where $\overline{[\rho(r)]}$ is the closure of the chamber containing $\rho(r)$.
	\end{proposition}
	
	\begin{proof}
		Since $\R^n_\textrm{dist}$ is dense in $\R^n$ and $G$ acts by continuous maps which leave $\R^n_\textrm{dist}$ invariant as a set, it suffices to show that $\bigcup_{r\in R}[\rho(r)]$ is a complete set of orbit representatives for $G$ acting on $\R^n_\textrm{dist}$. In fact $G$ simply permutes the components of $\R^n_\textrm{dist}$ so it suffices to show that $\bigcup_{r\in R}\rho(r)$ is a complete set of orbit representatives for the induced action of $G$ on $C$. 
		
		But now, $\rho$ is a bijection which exhibits an equivalence between the action of $G$ on $C$ and the action of $G$ on $S_n$ so we just need to show that $R$ is a complete set of orbit representatives for $G$ acting on $S_n$. The orbits of this action are precisely the right cosets of $G$, which completes the proof.
	\end{proof}
	
	\subsection{Gallery connectedness and fundamental domains}
	
	Given a set of right coset representatives $R$ for $G\le S_n$, the interior of $\fdc$ as defined in the proposition will, in general, not be a fundamental domain because it will not be connected. We can reinterpret connectedness in terms of algebraic properties of $R$. First some geometric definitions.
	
	\begin{definition}
		Let $c,c'\in C$ be distinct, we say the chambers $[c]$ and $[c']$ are \emph{adjacent} if $\overline{[c]}\cap\overline{[c']}$ has codimension 1. A \emph{gallery} is a sequence of chambers $[c_1],\dots,[c_k]$ such that consecutive chambers are adjacent. A set of chambers is called \emph{gallery connected} if any two distinct chambers in the set can be connected by a gallery which is completely contained in the set. As a shorthand, we sometimes call a subset $C'\subset C$ gallery connected if the set $\{[c]\mid c\in C'\}$ is gallery connected.
	\end{definition}
	
	It turns out that the decomposition of $\R^n_{\textrm{dist}}$ into chambers corresponds to the chamber system of $S_n$ acting on its Coxeter complex, about which we will not elaborate here, but the interested reader should consult \cite{Brown1989}. The upshot of this viewpoint is two characterisations of adjacency of chambers.
	
	\begin{lemma}\label{lemma:adjacency}
		Let $c\neq c'\in C$ and define $s=\rho^{-1}(c)$, $s'=\rho^{-1}(c')$. Then the following are equivalent:
		\begin{enumerate}\itemsep0em
			\item\label{point:adj} The chambers $[c]$ and $[c']$ are adjacent
			\item\label{point:alg} There is $1\le j\le n-1$ such that $s'=s(j\;j+1)$ where $(j\;j+1)$ is a transposition in $S_n$
			\item\label{point:comb} The vectors $c$ and $c'$ differ by swapping exactly two entries which are consecutive integers
		\end{enumerate}
	\end{lemma}
	
	\begin{proof}
		The equivalence of (\labelcref{point:adj}) and (\labelcref{point:alg}) is proved in, for example, Theorem I.5A of \cite{Brown1989}. To see the equivalence of (\labelcref{point:alg}) and (\labelcref{point:comb}), notice that $\rho(s(j\;j+1))=((s(j\;j+1))^{-1}(i))_i=:(c'_i)_i$. For $i\not\in \{j,j+1\}$, $c'_i=s(i)=c_i$ (where $c\coloneqq(c_i)_i$), whereas $c'_j=c_{j+1}$ and $c'_{j+1}=c_j$.
	\end{proof}
	
	The equivalence of (\labelcref{point:adj}) and \labelcref{point:comb} for the example of $S_4$ can be seen in (\cref{fig:chambers}). We will use this characterisation to prove \cref{prop:chamber-connected} which is key to showing that the image of $\pi_\uparrow$ is connected. We are now in a position to upgrade \cref{prop:G-orbit-reps} so that it produces a fundamental domain for the action of $G$. To state this, we define a \emph{right transversal} of $G\le S_n$ to be a minimal set of right coset representatives (ie a set containing exactly one element from every right coset).
	
	\begin{proposition}\label{prop:G-fund-dom}
		Let $R\subset S_n$ be a right transversal for $G\le S_n$ such that $\rho(R)$ is gallery connected. Then $\fd$, the interior of $\bigcup_{r\in R}\overline{[\rho(r)]}$, is a fundamental domain for $G$ acting on $\R^n$.
	\end{proposition}
	
	\begin{proof}
		By the definition, if $[c]$ and $[c']$ are adjacent, then the interior of $\overline{[c]}\cup\overline{[c']}$ will be connected. By induction on the length of galleries in $\{[\rho(r)]\mid r\in R\}$ it follows that $\fd$ is connected. It is also open by definition. 
		
		By \cref{prop:G-orbit-reps} we know that $\fdc$ is a complete set of orbit representatives for $G$. Finally, suppose that some $G$-orbit meets $\fd$ in at least two points, say $x$ and $x'$ and $g\in G$ is such that $g\cdot x=x'$. Since the $G$-action permutes the chambers, there are two possibilities:
		\begin{enumerate}\itemsep0em
			\item There are coset representatives $r,r'\in R$ such that $x\in [\rho(r)]$ and $x'\in [\rho(r')]$.
			\item There are coset representatives $r_1\neq r_2,r'_1\neq r'_2\in R$ such that $x\in \overline{[\rho(r_1)]}\cap\overline{[\rho(r_2)]}$ and $x'\in \overline{[\rho(r_1')]}\cap\overline{[\rho(r_2')]}$.
		\end{enumerate}
		
		In the first case we must have that $g\cdot[\rho(r)]=[\rho(r')]$, in which case it follows from \cref{lemma:Sn-fund-dom} and the fact that $g\neq 1$, that $r\neq r'$. But then by the equivalence of the action with the action on $S_n$, we have that $g\cdot r=gr=r'$ and $r$ and $r'$ represent the same right coset of $G$. This contradicts the assumption that $R$ is minimal. In the second case we can similarly argue that $\{r_1,r_2\}\neq\{r_1',r_2'\}$ but $g\cdot \{r_1,r_2\}=\{r_1',r_2'\}$, again contradicting the minimality of $R$. In either case $g$ cannot exist.
	\end{proof}
	
	\subsection{An algorithm to find coset representatives} \label{subsec:coset-reps}
	
	In this section we will summarise the main construction of \cite{Dixon1988} which gives an efficient algorithm to compute a right transversal for an arbitrary subgroup $G\le S_n$. The first step, as it is to find a base $B\subset N$ for $G\le S_n$. Set $B_0=()$, the empty tuple. We will assume that we have already constructed $B_{i-1}$ and computed $G_{i-1}$. If $G_{i-1}=\{1\}$, $B=B_{i-1}$ is a base and we are done. Otherwise, pick $b_i\in N$ such that $G_{i-1}\not\le\Stab_G(b_i)$ (it is easy to check that such a $b_i$ will always exist). Then let $B_i$ be $B_{i-1}$ with $b_i$ appended.
	\par
	Let $B=(b_1,\dots,b_k)$ be a base and recall we define $G_0=G$ and $G_i=G_{i-1}\cap\Stab_G(b_i)$ for $1\le i\le k$. We also write $\Delta_i=b_i\cdot G_{i-1}$ for the orbit of $b_i$ under $G_{i-1}$. Recursively construct a partition $\Pi_i$ of $N$, starting with $\Pi_0=\{N\}$. Denote by $\Gamma_i$ the element of $\Pi_{i-1}$ which contains $b_i$. One can check by induction that $\Gamma_i$ contains $\Delta_i$ as a subset. Define $\Pi_i$ by replacing $\Gamma_i$ in $\Pi_{i-1}$ by the non-empty subsets from the list: $\{b_i\}$, $\Delta_i-\{b_i\}$, and $\Gamma_i-\Delta_i$.
	
	Now let $U_i$ be a right transversal for the group $\Sym(\Delta_i)\times\Sym(\Gamma_i-\Delta_i)$ in $\Sym(\Gamma_i)$, where $\Sym(\Omega)$ is the group of permutations of the set $\Omega$ (in the next section we will fix a particular choice for $U_i$), and finally let 
	\[
	H_i=\prod_{\Gamma\in\Pi_i}\Sym(\Gamma).
	\]
	Then define $R=H_kU_kU_{k-1}\cdots U_1$, where for subsets $A,B\subset S_n$, $AB\coloneqq\{ab\mid a\in A,\;b\in B\}$.
	
	\begin{theorem}[\cite{Dixon1988} $\S 4$]\label{thm:right-trans}
		The set $R$ is a right transversal for $G\le S_n$.
	\end{theorem}
	
	\subsection{Gallery connected sets of coset representatives}
	
	We will now show how the method described above can be used to construct a right transversal $R$ for $G$ such that $\rho(R)$ is gallery connected. This is done by choosing a suitable base $B$, possibly re-indexing the set $N$, and choosing appropriate right transversals $U_i$ for $\Sym(\Delta_i)\times\Sym(\Gamma_i-\Delta_i)$ in $\Sym(\Gamma_i)$. We will prove \cref{thm:comb-alg} assuming $B$ and $N$ have been chosen in this way, and then in \cref{subsec:finish-proof} show that the assumptions on $B$ and $N$ can be dropped. 
	
	We described how to find a base for $G$ by appending more and more elements of $N$ to $B=()$ until $G_k=\{1\}$ in \label{subsec:coset-reps}. The first assumption we make is that each new $b_i$ is minimal in the orbit $b_i\cdot G_{i-1}$ with respect to the normal ordering on $N$. We will call such a base \emph{orbit minimal}. We will use the following lemma to build gallery connected sets out of other gallery connected sets.
	
	\begin{lemma}\label{lem:chamberConnected}
		Let $A_1,\dots,A_l$ be subsets of $S_n$ so that each contains the identity permutation $(1)$, and $\rho(A_i)$ is gallery connected for each $i$. Then $\rho(A_1A_2\cdots A_l)$ is gallery connected.
	\end{lemma}
	
\begin{wrapfigure}[13]{r}{45mm}
		\vspace{-5mm}
		\centering
		\begin{tikzpicture}[scale=1.2]
			\filldraw[thick, rounded corners=10, fill=cyan!25!white, draw=cyan] (0,0) rectangle (3,4);
			\node[cyan, rotate=90] at (0,2) [anchor=south]{\footnotesize$\rho(A_1A_2)$};
			
			\filldraw[thick, rounded corners=8, fill=black!25!white, draw=black, opacity=.5] (0.1,0.1) rectangle (2.9,1);
			\node at (1,1) [anchor=south]{\footnotesize$\rho(A_2)$};
			\filldraw[thick, rounded corners=9, fill=black!25!white, draw=black, opacity=.5] (2.95,0.05) rectangle (2,3.95);
			\node[rotate=90] at (2,2) [anchor=south]{\footnotesize$\rho(A_1)$};
			
			\filldraw[thick, rounded corners=8, fill=gray!25!white, draw=gray, opacity=.5] (0.1,3) rectangle (2.9,3.9);
			\node at (1,3) [anchor=north]{\footnotesize$a_1\cdot\rho(A_2)$};
			
			\draw[ultra thick, magenta] (.6,.5) .. controls (1,.8) .. (1.8,.4) .. controls (2.25,.2) .. (2.5,.5);
			\node[magenta] at (1.8,.6) [anchor=north east]{\footnotesize$p_{a_2}$};
			\draw[ultra thick, magenta] (.6,3.4) .. controls (1,3.7) .. (1.8,3.3) .. controls (2.25,3.1) .. (2.5,3.4);
			\node[magenta, rotate=-18] at (1.5,3.65) {\footnotesize$a_1\cdot p_{a_2}$};
			
			\draw[ultra thick, magenta] plot [smooth] coordinates {(2.5,.5) (2.5,.7) (2.8,1.3) (2.1,2) (2.4,2.4) (2.7,2.8) (2.5,3.4)};
			\node[magenta] at (2.1,2) [anchor=west]{\footnotesize$p_{a_1}$};
			
			\fill[magenta!70!black] (.6,.5) circle (2pt) node [anchor=north]{\small$a_2$};
			\fill[magenta!70!black] (2.5,.5) circle (2pt) node at (2.55,.4) [anchor=south east]{\small$(1)$};
			\fill[magenta!70!black] (2.5,3.4) circle (2pt) node [anchor=south]{\small$a_1$};
			\fill[magenta!70!black] (.6,3.4) circle (2pt) node [anchor=north]{\small$a_1a_2$};
		\end{tikzpicture}
		\vspace{-1mm}
		\caption{Building a gallery in $\rho(A_1A_2)$.}	
		\label{figure:gallery}	
	\end{wrapfigure}
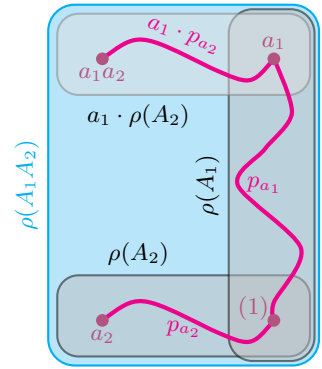

	\begin{proof}
		Notice that $A_1A_2$ contains $(1)(1)=(1)$. Let $a\in A_1A_2$, and choose $a_1\in A_1$ and $a_2\in A_2$ such that $a=a_1a_2$. Since $\rho(A_1)$ and $\rho(A_2)$ are gallery connected, there are galleries $p_{a_1}\subset \rho(A_1)$ and $p_{a_2}\subset \rho(A_2)$ which connect $\rho((1))$ to $\rho(a_1)$ and $\rho((1))$ to $\rho(a_2)$ respectively. Then $a_1\cdot p_{a_2}$ connects $\rho(a_1)$ to $\rho(a_1a_2)$ in $\rho(a_1A_2)$, and the concatenation $p_{a_1}\ast (a_1\cdot p_{a_2})$ is a gallery which connects $\rho((1))$ to $\rho(a)$ in $\rho(A_1A_2)$. Call this gallery $\tilde{p}_a$, its construction is illustrated in \cref{figure:gallery}. Now given $a,a'$ in $A_1,A_2$ the gallery $\tilde{p}_a^{-1}\cup\tilde{p}_{a'}$ (where $\tilde{p}_a^{-1}$ indicates $\tilde{p}_a$ traversed in reverse) connects $\rho(a)$ to $\rho(a')$ in $\rho(A_1A_2)$, so $\rho(A_1A_2)$ is gallery connected. The claim now follows by induction on $l$.
	\end{proof}

	From the definition of $R$ in the previous section, if we show that $H_k$ and each of the $U_i$'s satisfy the hypotheses of this lemma, then it will follow that $R$ is gallery connected. We will first consider $H_k$. Notice that in fact, if $\Pi_k=\{N_1,\dots,N_l\}$ then
	\[
	H_k=\prod_{i=1}^l\Sym(N_i)=\Sym(N_1)\Sym(N_2)\cdots\Sym(N_l)
	\] 
	can be written as a product of sets, again as in the lemma. Each $\Sym(N_i)$ contains $(1)$, so we just need $\rho(\Sym(N_i))$ to be gallery connected for each $i$. It follows immediately from \cref{lemma:adjacency} that this is the case if and only if $N_i$ is a sequence of consecutive digits from $N$. 
	
	In general this will not be the case, however it can be readily achieved by re-indexing the set $N$. In fact we can do this so that each part of each partition $\Pi_i$ is a set of consecutive digits, which will subsequently aid in showing that $\rho(U_i)$ is gallery connected.
	
	\begin{lemma}\label{lemma:re-indexing}
		We can re-index $N$ so that $b_i$ remains minimal in $\Delta_i$ and each part of $\Pi_i$ is a set of consecutive digits for $1\le i \le k$.
	\end{lemma}
	
	\begin{proof}
		We do induction on $i$: note that in $\Pi_0=\{N\}$ the only part is a set of consecutive numbers. Assume that each element of $\Pi_{i-1}$ is a set of consecutive digits, in particular $\Gamma_i\in \Pi_{i-1}$ is a set of consecutive digits. Assume one of the three subsets $\{b_i\}$, $\Delta_i-\{b_i\}$ or $\Gamma_i-\Delta_i$ is non-empty and does not consist of consecutive digits, then by the minimality of $b_i$ both $\Delta_i-\{b_i\}$ and $\Gamma_i-\Delta_i$ must be non-empty and not consist of consecutive digits. Re-index the elements of $\Gamma_i$ so that overall the same set of digits is used, but now $b_i$ is the smallest, the next smallest digits are all in $\Delta_i-\{b_i\}$, and the remaining digits are in $\Gamma_i-\Delta_i$.
	\end{proof}
	
	\subsection{Choosing a right transversal $U_i$}
	
	Finally we want to choose the right transversals $U_i$ for $\Sym(\Delta_i)\times\Sym(\Gamma_i-\Delta_i)$ in $\Sym(\Gamma_i)$. Write $\Delta_i=\{d_1,d_2,\dots,d_m\}$, and $\Gamma_i-\Delta_i=\{d_{m+1},d_{m+2},\dots,d_{m+m'}\}$. For $0\le l\le \min\{m,m'\}$, choose $d_{j_1}<d_{j_2}<\dots<d_{j_l}$ and ${d_{m+j'_1}<d_{m+j'_2}<\cdots<d_{m+j'_l}}$, and consider the product of transpositions
	\begin{equation}\label{eq:transversal}
	(d_{j_1}\;d_{m+j'_1})(d_{j_2}\;d_{m+j'_2})\cdots(d_{j_l}\;d_{m+j'_l})\in \Sym(\Gamma_i).
	\end{equation}
	Define $\widetilde{U}_i$ to be the set of all such products for any choice of $l$, and indices $j_k$ and $j'_{k}$.
	
	\begin{lemma}(\cite{Dixon1988} Lemma 2)
		$\widetilde{U}_i$ is a right transversal for $\Sym(\Delta_i)\times\Sym(\Gamma_i-\Delta_i)$ in $\Sym(\Gamma_i)$.
	\end{lemma}
	
	Let $\tilde{u}\in \widetilde{U}_i$ have the form given in \cref{eq:transversal}. If $l=0$, then $\tilde{u}$ is the identity and thinking of it as an element of $S_n\ge\Sym(\Gamma_i)$, we get $\rho((1))=(1,\dots,n)$. More generally $\rho(\tilde{u})$ will be the result of swapping each of the pairs $d_{j_k}\leftrightarrow d_{m+j'_k}$ in this vector, for $1\le k\le l$. Let $g_{\tilde{u}}\in\Sym(\Delta_i)\times\Sym(\Gamma_i-\Delta_i)$ be the permutation such that $\rho(g_{\tilde{u}}\cdot\tilde{u})$ has its first $m$ entries in increasing order, and its last $m'$ entries in increasing order. Define $U_i=\{g_{\tilde{u}}\cdot\tilde{u}\mid \tilde{u}\in\widetilde{U}_i\}$.
	
	\begin{lemma}
		$U_i$ is a right transversal for $\Sym(\Delta_i)\times\Sym(\Gamma_i-\Delta_i)$ in $\Sym(\Gamma_i)$.
	\end{lemma}
	
	\begin{proof}
		$\widetilde{U}_i$ contains exactly one element from each right coset of $\Sym(\Delta_i)\times\Sym(\Gamma_i-\Delta_i)$ in $\Sym(\Gamma_i)$. For $\tilde{u}\in\widetilde{U}_i$, the element $g_{\tilde{u}}\cdot\tilde{u}=g_{\tilde{u}}\tilde{u}$ lies in the same right coset as $\tilde{u}$ since $g_{\tilde{u}}\in\Sym(\Delta_i)\times\Sym(\Gamma_i-\Delta_i)$. Hence $U_i$ contains exactly one element from each right coset of $\Sym(\Delta_i)\times\Sym(\Gamma_i-\Delta_i)$ in $\Sym(\Gamma_i)$.
	\end{proof}
	
	We want to show that $\rho(U_i)$ is gallery connected, and for that we will use the re-indexing of $N$ provided by \cref{lemma:re-indexing}. Recall that $b_i$ is the $i$th element of the base $B$; it follows from our construction of $U_i$ that
	\begin{align}\label{eq:rho(U_i)}
		\rho(U_i)=\{(1,\dots,&b_i-1,\overset{\textrm{indexed by}\;\Delta_i}{\overbrace{c_1,\dots,c_m}},\overset{\textrm{indexed by}\;\Gamma_i-\Delta_i}{\overbrace{c_{m+1},\dots,c_{m+m'}}},b_i+m+m',\dots,n)\mid\nonumber\\
		&c_j\in\Gamma_i=\{b_i,b_i+1,\dots,b_i+m+m'-1\}\;\textrm{for all}\;1\le j\le m+m',\nonumber\\
		&c_1<\cdots<c_m,\;\textrm{and}\;c_{m+1}<\cdots<c_{m+m'}\}
	\end{align}
	Since this is notationally rather cumbersome, we will abbreviate elements of $\rho(U_i)$ by 
	\[
	(c_1,\dots,c_m\mid c_{m+1},\dots,c_{m+m'}), 
	\]
	where the first `half' consists of entries indexed by $\Delta_i$, and the second `half' consists of entries indexed by $\Gamma_i-\Delta_i$. 
	
	\begin{proposition}\label{prop:chamber-connected}
		Let $G\le S_n$, $B$ be an orbit minimal base, and $N$ indexed so that each part of $\Pi_i$ is a set of consecutive digits. Then the set $\rho(U_i)$ is gallery connected.
	\end{proposition}
	
	\begin{proof}	
		To help simplify notation, we will not distinguish between points in $C$ and the chambers they represent. We will show that $\rho(U_i)$ is gallery connected by explicitly constructing a gallery which joins an arbitrary chamber 
		\[
		c=(c_1,\dots,c_m\mid c_{m+1},\dots,c_{m+m'})\in \rho(U_i)
		\]
		to the chamber corresponding to the identity in $U_i$,
		\[
		\hat{c}=\rho((1))=(b_i,\dots,b_i+m-1\mid b_i+m,\dots,b_i+m+m'-1).
		\]
		\Cref{lemma:adjacency} gives the condition for consecutive chambers in this gallery to be adjacent: they must differ by swapping two entries which are consecutive integers. Furthermore we will ensure this gallery remains in $\rho(U_i)$ throughout. This implies that after swapping the two entries, the two halves of $c$ must remain properly ordered. Taken together, this implies that:
		\begin{displayquote}
			\itshape The only swaps we can perform must switch the position of an entry in the left half with one in the right half, and these entries must be consecutive integers.
		\end{displayquote}
		
		Let $c\in \rho(U_i)$ be arbitrary, write $\hat{c}_j=b_i+j-1$ for the $j$th entry of $\hat{c}$, and define
		\[
		\de(c)=\left(\sum_{j=1}^{m}c_j-\hat{c}_j\right)-\left(\sum_{j=m+1}^{m+m'}c_j-\hat{c}_j\right)
		\]
		which measures the degree to which $c$ and $\hat{c}$ differ.
		
		\paragraph{Claim 1} $\de(c)\ge0$.
		
		\noindent Let $j\le m$, then since the entries in the left half of $c$ are ordered, distinct integers greater than or equal to $b_i$, ${c_j\ge b_i+(j-1)=\hat{c}_j}$ so each term in the first sum is non-negative. Similarly, for $j>m$ the entries in the right half of $c$ are ordered, distinct integers less than or equal to $b_i+m+m'-1$, so $c_j\le b_i+m+m'-1-(m+m'-j)=b_i+(j-1)=\hat{c}_j$ so each term in the second sum is non-positive. \qedsymbol
		
		As a remark, it follows from this claim that $\de$ equals the $L^1$ distance between $c$ and $\hat{c}$. We shall perform a sequence of swaps as described above which have the effect decreasing the value of $\de(c)$. Since $\de(c)=0$ implies that $c=\hat{c}$, the required gallery can be constructed by induction on $\de(c)$. Assume $c\neq \hat{c}$, and let $j$ be the minimal index such that $c_j\neq \hat{c}_j$. Since the two halves of $c$ are ordered, $c_j$ is in the left half. 
		
		\paragraph{Claim 2} $c_{j'}\coloneqq c_j-1$ is in the right half of $c$.
		
		\noindent Indeed suppose it is in the left half, then by the ordering on $c$, $j'<j$, and by the minimality of $j$, $c_{j'}=\hat{c}_{j'}=b_i+j'-1$. But then 
		\[
		\hat{c}_j\neq c_j=c_{j'}+1=b_i+(j'+1)-1=\hat{c}_{j'+1},
		\]
		so $j\neq j'+1$ since all entries of $\hat{c}$ are distinct. But now $c_{j'}<c_{j'+1}<c_j$ (by the ordering on $c$), which is contradiction since these entries are distinct integers, and $c_{j'}$ and $c_j$ differ by 1. \qedsymbol
		
		Thus, $c_j$ and $c_j-1$ are entries in different halves of $c$ which are consecutive integers. Let $c'$ be the result of swapping these two entries in $c$, then
		\[
		\de(c)-\de(c')=((c_j-\hat{c}_j)-(c_{j'}-\hat{c}_{j'}))-((c_{j'}-\hat{c}_j)-(c_{j}-\hat{c}_{j'}))=2(c_j-c_{j'})=2>0
		\]
		so performing the swap strictly decreases $\de$. By induction, there is a gallery in $\rho(U_i)$ joining $c$ and $\hat{c}$, and hence $\rho(U_i)$ is gallery connected.
	\end{proof}

	It follows directly from this proposition, \cref{thm:right-trans}, and \cref{prop:G-fund-dom} that $R$ as defined in \cref{subsec:coset-reps} corresponds to a fundamental domain.
	
	\begin{corollary}\label{cor:fund-dom}
		Let $G\le S_n$, $B$ be an orbit minimal base, and $N$ indexed so that each part of $\Pi_i$ is a set of consecutive digits. Let $R$ be the right transversal for $G$ constructed above, then $\fd$, the interior of $\bigcup_{r\in R}\overline{[\rho(r)]}$, is a fundamental domain for $G$ acting on $\R^n$.
	\end{corollary}
	
	\subsection{Finishing off the proof}\label{subsec:finish-proof}
		
	We have two things to do to complete the proof of \cref{thm:comb-alg}: first show that the map $\pi_\uparrow$ as defined in \cref{subsec:comb-projections} has image in $\fdc=\bigcup_{r\in R}\overline{[\rho(r)]}$, and therefore indeed projects onto a fundamental domain, and then remove the assumptions of orbit minimality and on how $N$ is indexed.
	
	\begin{proposition}\label{prop:pi-img}
		Let $G\le S_n$, $B$ be an orbit minimal base, and $N$ indexed so that each part of $\Pi_i$ is a set of consecutive digits. Then the image of $\pi_\uparrow$ lies in $\bigcup_{r\in R}\overline{[\rho(r)]}$.
	\end{proposition}

	\begin{proof}
		It suffices to show that the image of $\R^n_\textrm{dist}$ lies in $\fdc=\bigcup_{r\in R}\overline{[\rho(r)]}$. We claim that
		\[
		\rho(R)=\{(c_j)_j\in C\mid \textrm{for}\;1\le i\le k,\; c_{b_i}\le c_j\;\textrm{for all}\;j\in\Delta_i\}.
		\]
		It is clear that the definition of $\pi_\uparrow$ implies that the right hand side of this is the image of $\left.\pi_\uparrow\right|_C$, so the proposition follows immediately from this claim. 
		
		Call the set on the right hand side $C'$, first we will show that $\rho(R)\subseteq C'$. By \cref{eq:rho(U_i)} (note that the entries of $(c_j)_j$ there are indexed differently there) we can see
		\[
		\rho(U_i)\subset\{(c_j)_j\in C\mid c_{b_i}\le c_j\;\textrm{for all}\;j\in\Delta_i\}.
		\]
		Since $U_i\subset \Sym(\Gamma_i)$, which fixes $b_{i-1}$ for $i\ge 2$, one can inductively check from the definition of $\rho$ that $\rho(U_k\cdots U_1)\subset C'$. Similarly, in the partition $\Pi_k$, each $b_i$ appears as a singleton, so $H_k$ also fixes $b_i$ for $1\le i\le k$, hence $\rho(R)\subseteq C'$.
		
		To establish the claim we just need to show that $|C'|=|\rho(R)|$, since they are finite sets this will imply that they are equal as sets. On the one hand, since $\rho$ is a bijection, and using Lagrange's Theorem
		\[
		|\rho(R)|=|R|=|\{\textrm{right cosets of}\;G\;\textrm{in}\;S_n\}|=|S_n|/|G|.
		\]
		On the other hand, each condition `$c_{b_i}\le c_j\;\textrm{for all}\;j\in\Delta_i$' decreases the size of $C$ by a factor of $|\Delta_i|$, so  
		\[
		|C'|=\frac{|C|}{|\Delta_1|\cdots|\Delta_k|}.
		\] 
		Since $C$ is the bijective image of $S_n$ under $\rho$, $|C|=|S_n|$. By the Orbit-Stabiliser Theorem, we also have 
		\[
		|\Delta_i|=|b_i\cdot G_{i-1}|=|G_{i-1}|/|\Stab_{G_{i-1}}(b_i)|=|G_{i-1}|/|G_{i}|,
		\]
		Where the last equality follows from the definition $G_i=\Stab_{G_{i-1}}(b_i)$. Therefore
		\[
		|\Delta_1|\cdots|\Delta_k|=\frac{|G_{0}|}{|G_1|}\frac{|G_{1}|}{|G_2|}\cdots\frac{|G_{k-1}|}{|G_k|}=\frac{|G_0|}{|G_k|}=\frac{|G|}{|\{1\}|}=|G|
		\]
		Hence $|C'|=|\rho(R)|$, which completes the proof.
	\end{proof}

	\begin{proof}[Proof of \cref{thm:comb-alg}]
		Let $N=\{1,\dots,n\}$, and choose $B$ a base for $G\le S_n$, and  $\varepsilon $ satisfying the conditions in \cref{subsec:perturb}. Let $s\in S_n$ be a permutation of $N$ such that $B^s\coloneqq B\cdot s$ is an orbit minimal base, and each part of each partition $\Pi_i^s\coloneqq\Pi_i\cdot s$ is a set of consecutive digits.  That $s$ exists is clear by first permuting $k$ times so that $B$ is orbit minimal (note $b_i\not\in\Delta_j$ for all $j>i$) and then applying \cref{lemma:re-indexing}. Write $b_i^s=b_i\cdot s$ so that $B^s=(b_1^s,\dots,b_k^s)$.
		
		Let $G^s=s^{-1}Gs$ be the conjugate of $G$ by $s$ in $S_n$, then for any $g\in G$ and $m\in N$, 
		\begin{equation}\label{eq:conj-action}
		(m\cdot s)\cdot g^s=((g^s)^{-1}s^{-1})(m)=(s^{-1}g^{-1})(m)=(m\cdot g)\cdot s
		\end{equation}
		where $g^s=s^{-1}gs$. In other words, permuting by $s$ and then acting by $G^s$ is the same as acting by $G$ and then permuting by $s$. It follows that $G_i^s\coloneqq s^{-1}G_is=G^s_{i-1}\cap\Stab_{G^s}(b_i^s)$, and  $\Delta_i^s\coloneqq\Delta_i\cdot s=b_i^s\cdot G_{i-1}^s$. 
		
		Finally define $\phi^s_\uparrow$ and $\pi_\uparrow^s$ as in \cref{subsec:comb-projections} with respect to $B^s$ and $\varepsilon $. We claim that for $\hat{x}$ as defined in \cref{subsec:perturb}, $\phi^s_\uparrow(\hat{x})=\left(\phi_\uparrow(\hat{x})\right)^s=g_{\hat{x}}^s$. Indeed by definition $\phi^s_\uparrow(\hat{x})=\tilde{g}_k\cdots \tilde{g}_1$ where $\tilde{g}_i\in G^s_i$ such that $\tilde{\jmath}\cdot \tilde{g}_i=b^s_i$ and $\tilde{\jmath}\in\Delta^s_i$ is chosen such that the $\tilde{\jmath}$th entry of $(\tilde{g}_{i-1}\cdots \tilde{g}_1)\cdot\hat{x}$ is minimal among those entries indexed by $\Delta^s_i$. But now $\Delta^s_i=\Delta_i\cdot s$ means $\tilde{\jmath}=j\cdot s$ (where $j\in\Delta_i$ is the index found in the definition of $\phi_\uparrow$). Thus 
		\[
		b^s_i=b_i\cdot s=(j\cdot g_i)\cdot s\overset{\textnormal{\cref{eq:conj-action}}}{=}(j\cdot s)\cdot g_i^s=\tilde{\jmath}\cdot  g_i^s,
		\]
		so we can certainly choose $\tilde{g}_i=g_i^s$. Then as claimed
		\[
		\phi^s_\uparrow(\hat{x})=\tilde{g}_k\cdots \tilde{g}_1=g_k^s\cdots g_1^s=(g_k\cdots g_1)^s=g_{\hat{x}}^s=(\phi_\uparrow(\hat{x}))^s.
		\]
		Expanding out $\phi^s(\hat{x})=s^{-1}\phi(\hat{x})s$, we can now compute $\pi^s_\uparrow$ in terms of $\pi_\uparrow$ and $s$:
		\[
		\pi^s_\uparrow(x)=\phi^s_\uparrow(\hat{x})\cdot x=s\cdot(\phi_\uparrow(\hat{x})\cdot(s^{-1}\cdot x))=s\cdot\pi_\uparrow(s^{-1}\cdot x).
		\]
		Writing $\fd$ for the interior of the image of $\pi_\uparrow$, and $\fd^s$ for the interior of the image of $\pi^s_\uparrow$, this implies $\fd^s=s\cdot\fd$ (because $s^{-1}\cdot\R^n=\R^n$). But \cref{cor:fund-dom} together with \cref{prop:pi-img} says that $\fd^s$ is a fundamental domain for $G^s$ and $\pi^s_\uparrow$ is a projection onto $\fd^s$; so $\fd=s^{-1}\cdot\fd^s$ is a fundamental domain for $G$ and $\pi_\uparrow$ is a projection onto $\fd$.
		
		To prove the final claim of the theorem, that $\pi_\uparrow\colon\R^n\to\R^n$ is uniquely defined by the choice of $B$ and $\varepsilon $, we just need to show that a different choice of the elements $g_1$, \dots, $g_k$ given $x\in \R^n$ does not change $\phi_\uparrow$. In fact $\phi_\uparrow$ is determined completely by what it does to the points $\hat{x}\in\R^n_\textrm{dist}$, and $\phi_\uparrow(\hat{x})$ will lie inside the fundamental domain (not on its boundary). By (\labelcref{point:nonintersecting}) in the definition of a fundamental domain, any different choice $g'_1$, \dots, $g'_k$ must necessarily combine to give the same element $g_{\hat{x}}$ (no non-trivial element of $G$ acts trivially), and hence $\phi_\uparrow$ is uniquely determined.
	\end{proof}

	\section{Other mathematical results}\label{sec:other-proofs}
	
	\subsection{Universal approximation theorem}\label{subsec:univ-approx-thm}
	The universal approximation theorem is a fundamental result in the theory of machine learning that any continuous function $\al\colon X\to \R^m$ on a compact subset $X\subset \R^n$ can be arbitrarily well approximated by a neural network with one hidden layer. To state the theorem precisely, one needs to specify that closeness between two continuous functions $\al,\be\colon X\to \R^m$ is measured by their uniform distance, ie by
	\[
	\|\al-\be\|_{\infty}=\mathrm{sup}_{x\in X} \|\al(x)-\be(x)\|.
	\]
	Denote by $C(X,\R^m)$ the set of continuous functions $X\to \R^m$ and by $\mathfrak{M}_{n,m}(\sigma)$ the set of functions $\R^n\to \R^m$ implemented by a neural network with activation function $\sigma\colon \R\to \R$, one hidden layer with arbitrarily many neurons and $m$ output neurons. The universal approximation theorem, as in Theorem 2 of \cite{Hornik1991}, states that 	if $\sigma$ is continuous, bounded, and non-constant, then $\mathfrak{M}_{n,m}(\sigma)$ is dense in $C(X,\R^m)$ for all compact subsets $X$ of $\R^n$.
	
	To obtain a $G$-invariant version of this theorem, let $C^G(X,\R^m)$ be the set of continuous functions on $X$ that are $G$-invariant and let $\mathfrak{M}_{n,m}^G(\sigma)$ be the set of functions of the form $\al\circ \pi$ where $\pi\colon X\to \R^l$ is a $G$-invariant map which does not identify points in distinct $G$-orbits, for some $l$, and $\al \in \mathfrak{M}_{l,m}(\sigma)$.
	
	\begin{theorem}[$G$-invariant universal approximation theorem]
		If $\sigma$ is continuous, bounded and non-constant, then $\mathfrak{M}_{n,m}^G(\sigma)$ is dense in $C^G(X,\R^m)$ for any compact subset $X\subset \R^n$.
	\end{theorem}

	\begin{proof}
		We need to show that given $\varepsilon>0$, a $G$-invariant map $\al\colon X \to \R^m$, and another $G$-invariant map $\pi\colon X\to\R^l$, there is a neural network $\overline{\be}\in \mathfrak{M}_{l,m}(\sigma)$ such that $\|\al-\overline{\be}\circ \pi\|_{\infty}<\varepsilon$. Let $\overline{\alpha}\colon\R^l\to\R^m$ be the map induced from $\alpha$ by $\pi$ (here we need the fact that $\pi$ does not identify points in distinct $G$-orbits for $\overline{\alpha}$ to be well-defined). By the universal approximation theorem (Theorem 2 in \cite{Hornik1991}), there is a network $\overline{\be}\colon \R^l \to \R^m$ such that
		\[\|\overline{\alpha}-\overline{\be}\|_{\infty}<\epsilon.\]
		Furthermore, since $\al$ is $G$-invariant, $\al=\overline{\al}\circ \pi$, which implies that 
		\begin{align*}
			\|\al-\overline{\be}\circ \pi\|_{\infty}&=\|\overline{\al}\circ \pi-\overline{\be}\circ \pi\|_{\infty}=\|(\overline{\al}-\overline{\be})\circ \pi\|_{\infty}\leq \|\overline{\al}-\overline{\be}\|_{\infty}<\varepsilon.\qedhere
		\end{align*}
	\end{proof}
	Note that the same argument can be applied when considering $L^p$ norms or weighted Sobolev norms instead of the supremum norm, leading to analogous versions of Theorem 1, 3, and 4 in \cite{Hornik1991}. A projection onto a fundamental domain certainly satisfies the hypothesis on $\pi$.
	
	\subsection{Computing the space of linear equivariant maps}\label{subsec:comp-lin-equiv}
	
	We present a method to compute the space of equivariant linear maps as outlined in \cref{subsec-equiv-layers}.
	This can be a hard problem, however in the context of that section, the problem is transformed into the following: find all compatible maps $\overline{\lambda}\colon \R^{n_1}/G\to\R^{n_2}/G$ which lift to a linear map.
	On the face of it this is no simpler. However, if $G\le S_n$ acts by permutations on $\R^n$, then the span of $\mathds{1}$ is $G$-invariant so we get a decomposition $\R^n=\R\mathds{1}\oplus\mathds{1}^\perp$ where $G$ acts trivially on the first factor, and the second factor is the subspace orthogonal to $\mathds{1}$, which is also $G$-invariant. Moreover $G$ acts orthogonally on $\mathds{1}^\perp\cong\R^{n-1}$, and so the unit sphere $\mbS^{n-2}\subset\R^{n-1}$ is also $G$-invariant. The action of $G$ on $\mbS^{n-2}$ completely determines the action on $\R^n$ and the quotient space is, as a topological manifold,
	\begin{equation}\label{eq:inv-subspaces}
	\R^n/G=\R\mathds{1}\oplus\Cone(\mbS^{n-1}/G).
	\end{equation}
	Furthermore, $\mbS^{n-2}$ can be given the structure of a simplicial complex such that $G$ acts by simplicial isomorphisms. The simplicial structure is the first barycentric subdivision of the boundary of an $(n-1)$--simplex. This induces a simplicial structure on $\mbS^{n-2}/G$ (which is hinted at by \cref{lemma:Sn-fund-dom}: the chambers are the maximal simplices in $\mbS^{n-2}$; and illustrated in \cref{fig:chambers}). One can use this to reduce the problem of checking compatibility to a combinatorial one, since the stabiliser of a point $x\in\mbS^{n-1}$ (and hence of any point in $\R^n$) depends only on the unique simplex which contains $x$ in its interior. This approach can help compute the space of $G$-equivariant linear maps in general.

	\section{Implementation}\label{sec:implementation}
	
	\subsection{Machine Learning Experiments}
	
	All neural networks were trained using Keras.
	SVMs were fitted using sci-kit learn. In each case the performance was averaged over 10 runs, and sample standard deviations are given in the tables in \cref{sec:results}.
    
    \paragraph{Cayley tables}
    
    There are $5$ isomorphism classes of groups with $8$ elements:
    $C_8$, $C_4 \times C_2$, $D_4$, $Q_8$, $C_2 \times C_2 \times C_2$.
    We generated a dataset of $40000$ Cayley tables:
    a sixth of the tables were permutations of the group $C_8$, another sixth of the group $C_4 \times C_2$, another sixth of the group $D_4$, a quarter of the group $Q_8$, a quarter of the group $C_2 \times C_2 \times C_2$.
    The dataset was split into training and test dataset of equal size.
    Two models were compared:
    a fully connected neural network with two hidden layers of size $100$ and $10$ with activation function $\ReLU$ using the Adam optimiser with learning rate $0.001$ and cross-entropy loss as a loss function training for 200 epochs;
    and a linear SVM.
    
    \paragraph{CICY}
    
    Neural network architectures and training parameters were taken for the entries MLP, $\textrm{MLP}+\textrm{pre-processing}$, and Inception from \cref{table:cicy-results} were taken from the given references.
	The only difference we found is that in the case $\ord+$Inception on the randomly permuted dataset, accuracy increased when \emph{not} removing outliers from the training data, and we therefore decided to not remove outliers from the training data in this case, contrary to what was done in \cite{Erbin2021}.
	The group invariant neural network had $6$ equivariant layers (with $4$ trainable parameters each) with $100$ channels and cross-channel interactions, followed by sum pooling, and two fully connected layers with $64$ and $32$ neurons.
	No dropout was used.
	We experimented with max pooling, dropout, different numbers and sizes of layers and found the above parameters to work best.
	We found that test accuracy varied strongly for large batch sizes and eventually trained with batch size $1$.
	We randomly split the dataset into training and test sets of equal size.
    
    \paragraph{Classifying rotated handwritten digits}
    
    Two neural networks were compared:
    a fully connected neural network with no hidden layers; 
    and a fully connected neural network with two hidden layers of size $128$ and $64$ with activation function $\ReLU$.
    Using the Adam optimiser with learning rate $0.001$ and cross-entropy loss as a loss function we trained for 100 epochs. 

	\subsection{Approximating Dirichlet projections} \label{subsec:implementation-grad-desc}
	
	The map $\phi$ from \cref{subsec:grad-desc} maps $x\in \R^n$ to the element in $G$ which minimises the function $N(g\cdot x)$, where
	\[
	N\colon \R^n\to \R,\quad x \mapsto \langle  x,\hat{x}\rangle
	\]
	and $\hat{x}$ is a fixed element in $\R^n$.
	Our main application of the discrete gradient descent algorithm is for CICY matrices when $G=S_{12}\times S_{15}$. Since $|G|\approx 6\times 10^{20}$ one cannot minimise a function over $G$ by simply evaluating it at all elements of $G$. \par
	Instead, it is natural to compute the minimiser of $N$ on a group orbit using gradient descent. However, steps in the descent are restricted to the $G$-orbit of a point, which is a discrete set, so it is not clear how to define a gradient. We can circumvent this using the algebra of the group $G$. A \emph{generating set} $T$ for $G$ is a subset such that any element $g\in G$ can be written as a finite product of elements in $T$, ie there are $t_1,\dots, t_l \in T$ such that $g=t_1\cdots t_l$. Two points $x,x'\in \R^n$ are \emph{adjacent} with respect to $T$ if there is $t\in T$ such that $x'=t\cdot x$, so in particular, adjacent points are in the same $G$-orbit. 
	
	\begin{definition}
	 Given an action of a finite group $G$ on $\R^n$, a generating set $T$ of $G$, a function $N\colon \R^n\to \R$ and $x\in \R^n$, the \emph{discrete gradient descent} is an approximation for 
	 \[\min_{g\in G}N(g\cdot x)\]
	 and is defined iteratively as follows.
	 Let $x_0=x$. Given $x_i$, define 
	 \[x_{i+1}=\min_{t \in T\cup\{e\}} N(t \cdot x_i).\]
	 The output of the algorithm is $x_i$ when $x_{i+1}=x_i$.
	\end{definition}

	Since we can get between any two points in $G$ by a finite sequence of steps by generators this algorithm always terminates.
    In general, there are many choices for a generating set $T$ resulting in different approximations for $\phi$.
    For $G=S_n$ a  natural choice for a generating set is given by $T=\{(1\: 2),(2\: 3),\dots, (n-1\: n)\}$. In particular, one has in this case $|T|=n-1\ll n!=|S_n|$. By taking the union of these generating sets for $S_n$ and $S_m$ one obtains a generating set of size $n+m-2$ for $S_n\times S_m$.
    
	Choosing a larger generating set increases the computational cost of the algorithm but potentially also its accuracy. For example, consider the set $T'=\{tt' \mid t,t' \in T\cup\{e\} \}$. This is a generating set for $S_n$ and again yields a generating set for $S_n \times S_m$. When applied to the CICY dataset, choosing $T'$ instead of $T$ has led to a significant increase in computation cost, but not so in accuracy. Instead, we have used discrete gradient with different seeds.
	
	For a $12\times 15$ CICY matrix $x$, The seeds are $x_{km}\coloneqq{C_{12}}^k x {C_{15}}^m$, where $C_i$ are cyclic permutation matrices $1\leq k \leq 12$ and $1 \leq m \leq 15$. To each $x_{km}$, apply the discrete gradient descent algorithm above and pick the minimum of all seeds. The result is still a permutation of $x$ since $C_i$ are permutation matrices. This increases the computation cost by a constant factor $11\times 14+1=155$ but has led to a significant accuracy boost.


\end{document}